\documentclass[11pt,letterpaper]{article}
\usepackage{graphicx}
\usepackage{epsfig,subfigure}
\usepackage{amssymb}
\usepackage{amsthm}
\usepackage{amsfonts}
\usepackage{amsmath}
\usepackage[margin=1.4in]{geometry}
\RequirePackage{natbib}
\RequirePackage[colorlinks,citecolor=blue,urlcolor=blue]{hyperref}

\usepackage{array}
\usepackage{bm}
\usepackage{algorithm}
\usepackage{algorithmic}

\usepackage{rotating}
\usepackage{multirow}

\DeclareMathOperator*{\sign}{sign}

\newcommand{\norm}[1]{||#1||}
\newcommand{\opnorm}[2]{| \! | \! | #1 | \! | \! |_{{#2}}}
\DeclareMathOperator*{\ind}{1{\hskip -2.5 pt}\hbox{I}}

\newcommand{\iidsim}{\stackrel{{iid}}{\sim}}

\newcommand{\xv}{\mathbf{x}}
\newcommand{\Xv}{\mathbf{X}}
\newcommand{\yv}{\mathbf{y}}
\newcommand{\rv}{\mathbf{r}}
\newcommand{\ev}{\mathbf{e}}
\newcommand{\Qv}{\mathbf{Q}}
\newcommand{\Wv}{\mathbf{W}}
\newcommand{\Dv}{\mathbf{D}}
\newcommand{\thetav}{\bm{\theta}}
\newcommand{\etav}{\bm{\eta}}
\newcommand{\Sigmav}{\bm{\Sigma}}
\newcommand{\Deltav}{\bm{\Delta}}

\newcommand{\Pm}{\mathbb{P}}
\usepackage{color}

\theoremstyle{plain}
\newtheorem{thm}{Theorem}
\newtheorem{lem}[thm]{Lemma}
\newtheorem{prop}[thm]{Proposition}
\newtheorem{cor}[thm]{Corollary}

\numberwithin{equation}{section}
\theoremstyle{plain}

\title{\huge Sparsistent Estimation Of Time-Varying Markov Random Fields}
\date{}

\author{
Mladen Kolar ~ and ~ Eric P.~Xing\thanks{Machine Learning Department, Carnegie Mellon University, Pittsburgh, PA 15217, USA; e-mail: {\tt
$\{$mladenk,epxing$\}$@cs.cmu.edu}.}
}

\bibpunct{(}{)}{,}{a}{,}{,}

\begin{document}

\maketitle

\begin{abstract}
Network models have been popular for modeling and representing complex
relationships and dependencies between observed variables. When data
comes from a dynamic stochastic process, a single static network model
cannot adequately capture transient dependencies, such as, gene
regulatory dependencies throughout a developmental cycle of an
organism. Kolar et al (2010b) proposed a method based on
kernel-smoothing l1-penalized logistic regression for estimating
time-varying networks from nodal observations collected from a
time-series of observational data. In this paper, we establish
conditions under which the proposed method consistently recovers the
structure of a time-varying network. This work complements previous
empirical findings by providing sound theoretical guarantees for the
proposed estimation procedure. For completeness, we include numerical
simulations in the paper.
\end{abstract}


\section{Introduction}
\label{intro}

In recent years, we have witnessed fast advancement of
data-acquisition techniques in many areas, including biological
domains, engineering and social sciences. As a result, new statistical
and machine learning techniques are needed to help us develop a better
understanding of complexities underlying large, noisy data
sets. Networks have been commonly used to abstract noisy data and
provide an insight into regularities and dependencies between observed
variables. For example, in a biological study, nodes of the network
can represent genes in one organism and edges can represent
associations or regulatory dependencies among genes. In a social
domain, nodes of a network can represent actors and edges can
represent interactions between actors. Recent popular techniques for
modeling and exploring networks are based on the structure estimation
in the probabilistic graphical models, specifically, Markov Random
Fields (MRFs). These models represent conditional independence between
variables, which are represented as nodes. Once the structure of the
MRF is estimated, the network is drawn by connecting variables that
are conditionally dependent.

Classical literature mainly focuses on estimating a single static
network underlying a complex system. However, in reality, many systems
are inherently dynamic and can be better explained by a dynamic
network whose structure evolves over time. Consider the following real
world problems:\par
\begin{itemize}
\item{\it Analysis of gene regulatory networks.} Suppose that we have
  a set of $n$ microarray measurements of gene expression levels,
  obtained at different stages during the development of an organism
  or at different times during the cell cycle. Given this data,
  biologists would like to get insight into dynamic relationships
  between different genes and how these relations change at different
  stages of development. The problem is that at each time point there
  is only one or at most a few measurements of the gene expressions;
  and a naive approach to estimating the gene regulatory network,
  which uses only the data at the time point in question to infer the
  network, would fail. To obtain a good estimate of the regulatory
  network at any time point, we need to leverage the data collected at
  other time points and extract some information from them.\par
\item {\it Analysis of stock market.} In a finance setting, we have
  values of different stocks at each time point. Suppose, for
  simplicity, that we only measure whether the value of a particular
  stock is going up or down. We would like to find the underlying
  transient relational patterns between different stocks from these
  measurements and get insight into how  these patterns change over
  time. Again, we only have one measurement at each time point and we
  need to leverage information from the data obtained at nearby time
  points.\par
\item{\it Understanding social networks.} There are 100 Senators in
  the U.S. Senate and each can cast a vote on different bills. Suppose
  that we are given $n$ voting records over some period of time. How
  can one infer the latent political liaisons and coalitions among
  different senators and the way these relationships change with
  respect to time and with respect to different issues raised in bills
  just from the voting records?
\end{itemize}

The aforementioned problems have commonality in estimating a sequence
of time-specific latent relational structures between a fixed set of
entities (i.e., variables), from a time series of observation data of
entities states; and the relational structures between the entities
are time evolving, rather than being invariant throughout the data
collection period as commonly assumed in much of the current
literature on high-dimensional undirected network estimation (see,
e.g., \citet{meinshausen06high, bresler07reconstruction, yuan07model,
  banerjee07model, rothman08spice, friedman08sparse, ravikumar08high,
  fan09network, peng09partial, ravikumar09high, wang09learning,
  guo10joint} and references therein). Typically, the available data
for the problem are very scarce, with only one or at most a few
measurements corresponding to any particular latent structure, while
the total number of potential relations is large and exceeds the total
number of observations. However, as we will show later, under the
assumption that the network is sparse and slowly changes with time, it
is possible to consistently estimate the network structure at any time
point.
 
A popular model for the relational structure over a fixed set of
entities that is widely studied is the Markov random field
(MRF)~\citep{wainwright08graphical, getoor07introduction}.  Let
$G=(V,E)$ represent a graph, of which $V$ denotes the set of vertices,
and $E$ denotes the set of edges over vertices. Depending on the
specific application of interest, a node $u \in V$ can represent a
gene, a stock, or a social actor, and an edge $(u,v) \in E$ can
represent a relationship (e.g., correlation, influence, friendship)
between actors $u$ and $v$. Let $\Xv = (X_1,\ldots, X_p)'$, where
$p=|V|$, be a random vector of nodal states following a probability
distribution indexed by $\thetav \in \Theta$. Under a MRF, the nodal
states $X_u$'s are assumed to be discrete, i.e., $X_u \in {\cal
  X}\equiv \{s_1, \ldots, s_k\}$, and the edge set $E \subseteq V
\times V$ encodes certain conditional independence assumptions among
components of the random vector $\Xv$, for example, the random
variable $X_u$ is conditionally independent of the random variable
$X_v$ given the rest of the variables if $(u,v) \not\in E$. Under the
special case of binary nodal states, e.g., $X_u \in {\cal X}\equiv
\{-1, 1\}$, and assuming pairwise potential weighted by $\theta_{uv}$
for all $(u,v) \in E$ and $\theta_{uv} = 0$ for all $(u,v) \not\in E$,
the joint probability of $\Xv=\xv$ can be expressed by a simple
exponential family model: $\Pm_{\thetav}(\xv) = \frac{1}{Z}\exp \{
\sum_{u < v} \theta_{uv} x_u x_v \}$, also known as the Ising model,
where $Z$ denotes the partition function that is usually intractable
to compute. A statistical challenge is to estimate the network
topology determined by the edge set $E$ from the observed data $\xv^i
\iidsim \Pm_{\thetav}$ $(i=1,\ldots,n)$ with $n \ll p$.

In this paper, we study the problem of estimating a sequence of
high-dimensional MRFs that slowly evolve over time from observational
data. Suppose that we are given the data $\{\xv^t \sim
\Pm_{\thetav^t}\}_{t \in {\cal T}_n}$, where ${\cal T}_n = \{1/n, 2/n,
\ldots, 1\}$ is the time index set, then our goal is to estimate the
sequence of graphs $\{G^t\}_{t \in {\cal T}_n}$ underlying each
observation $\xv^t \sim \Pm_{\thetav^t}$ in the time series. In order
to make this estimation problem feasible, we will have to assume that
the underlying probability distribution changes smoothly, which we
define precisely later. Estimating a sequence of graphs provides us
with insight into the dynamics of the relational changes underlying
data. A reader should observe that commonly used methods, which try to
estimate a single static graph $G$ from data $\{\xv^i\}_i$ assumed to
be {\it i.i.d.} from $\Pm_{\thetav}$, cannot provide insight into
dynamic aspect of the underlying relational structure.

The main contribution of this paper is to establish theoretical
guarantees for the estimation procedure for time-varying networks
proposed in \citet{kolar08estimating}. The estimation procedure is
based on temporally smoothed $\ell_1$-regularized logistic regression
formalism, which is detailed in Section~\ref{estimation}. An
application to real world data was given in \citep{le09keller},
where the procedure was used to infer the latent evolving regulatory
network underlying 588 genes across the life cycle of
\textit{Drosophila melanogaster} from microarray time course. Although
the true regulatory network is not known for this organism, the
procedure recovers a number of interactions that were previously
experimentally validated. Since in most real world problems the
ground truth is not known, we emphasize the importance of simulation
studies to evaluate the estimation procedure.

It is noteworthy that the problem of the graph structure estimation is
quite different from the problem of (value-) consistent estimation of
the unknown parameter $\thetav$ that indexes the distribution. In
general, the graph structure estimation requires a more stringent
assumptions on the underlying distribution and the parameter
values. For example, observe that a consistent estimator of $\thetav$
in the Euclidean distance does not guarantee a consistent estimation
of the graph structure, encoded by the non-zero patter of the
estimator. In the motivating problems that we started with, the main
goal is to understand the interactions between different actors. These
interactions are more easily interpreted by a domain expert than the
numerical values of the parameter vector $\thetav$ and have potential
to reveal more information about the underlying process of
interest. This is especially true in situations where there is little
or no domain knowledge and one is interested in obtaining casual,
preliminary information.

Due to its importance in number of domains, including systems biology,
finance and signal processing, a number of authors have proposed
algorithms for inferring time inhomogeneous networks, many of which
have appeared after the initial draft of this paper was communicated
\citep{kolar09sparsistent}. The literature can be divided into two
categories: estimation of directed graphical models and estimation of
undirected graphical models. Literature on estimating
time-inhomogeneous directed networks usually assumes a time-varying
vector auto-regressive model for observed data \citep[see, for
example,][]{punskaya2002bayesian, fujita2007time, rao2007inferring,
  Grzegorczyk09nonstationary, song09time, robinson09nonstationary,
  robinson2010learning, yi10constructing, lebre10statistical,
  husmeier10Intertime, dondelinger2010heterogeneous,
  Grzegorczyk2011Nonhomogeneous, grzegorczyk2011improvements,
  wang2011time, grzegorczyk2012non, grzegorczyk2012bayesian,
  dondelinger12nonhomogeneous, lebre2012nonhomogeneous}, a class of
models that can be represented in the formalism of time-inhomogeneous
Dynamic Bayesian Networks although not all authors use terminology
commonly used in the Dynamic Bayesian Networks literature. Markov
switching linear dynamical systems are another popular choice for
modeling non-stationary time series \citep[see, for
example,][]{andrieu2003efficient, Yoshida2005estimating,
  dobigeon2007joint, siracusa2009tractable, fox2011bayesian,
  jiang12bayesian}.  This body of work has focused on developing
flexible models capable of capturing different assumptions on the
underlying system, efficient algorithms and sampling schemes for
fitting these models. Although a lot of work has been done in this
area, little is known about finite sample and asymptotic properties
regarding the consistent recovery of the underlying networks
structures. Some asymptotic results are given in
\cite{song09time}. Due to the complexity of MCMC sampling procedures,
existing work does not handle well networks with hundreds of nodes,
which commonly arise in practice. Finally, the biggest difference from
our work is that the estimated networks are
directed. \citet{vogel09on} point our that undirected models
constitute the simplest class of models, whose understanding is
crucial for the study of directed models and models with both,
directed and undirected edges. \cite{talih2005structural} and
\cite{Xuan2007modeling} study estimation of time-varying Gaussian
graphical models in a Bayesian setting. \cite{talih2005structural} use
a reversible jump MCMC approach to estimate the time-varying variance
structure of the data. \cite{Xuan2007modeling} proposed an iterative
procedure to segment the time-series using the dynamic programming
approach developed by \cite{fearnhead2006exact} and fit a Gaussian
graphical model using the penalized maximum likelihood approach on
each segment. To the best of our knowledge, \citet{zhou08time} is the
first work that focuses on consistent estimation, in the Frobenius
norm, of covariance and concentration matrix under the assumption that
the time-varying Gaussian graphical model changes smoothly over time.
Network estimation consistency for this smoothly changing model is
established in \cite{kolar2011time}. Time-varying Gaussian graphical
models with abrupt changes in network structure were studied in
\citet{kolar10estimating}, where consistent network recovery is
established using a completely different proof technique. A related
problem is that of estimating conditional covariance matrices
\citep{yin10nonparametric,kolar10nonparametric}, where in place of
time, which is deterministic quantity, one has a random
quantity. Methods for estimating time-varying discrete Markov random
fields were given in \citet{amr09tesla} and \citet{kolar08estimating},
however, no results on the consistency of the network structure were
given. As we will see later, showing that a time-varying discrete
undirected network is consistently estimated is a much harder task
than showing the same result for time-varying Gaussian graphical
models.

This paper is organized as follows. Section~\ref{model} introduces the
network model. The estimation procedure is reviewed in
Section~\ref{estimation}. The conditions under which the estimation
procedure consistently recovers the network structure are stated in
Section~\ref{theory}, together with the main theoretical result. The
proof is outlined in Section~\ref{proof} with technical details
presented in the appendix. Simulation results are given in
Section~\ref{simulation}.

\section{The Model}
\label{model}

We are given a sequence of $n$ nodal states $\mathcal{D}_n = \{ \xv^t
\sim \mathbb{P}_{\thetav^t} | t \in \mathcal{T}_n\}$, with the time
index defined as $\mathcal{T}_n = \{1/n, 2/n, \ldots, 1\}$. For
simplicity of presentation, we will assume that the observations are
equidistant in time and only one observation is available at each time
point from distribution $\mathbb{P}_{\thetav^t}$ indexed by
$\thetav^t$.  Specifically, we assume that the $p$-dimensional random
vector $\Xv^t$ takes values in $\{-1, 1\}^p$ and the probability
distribution takes the following form:
%%%
\begin{equation} \label{eq:model_mrf_intro}
  \mathbb{P}_{\thetav^t}(x) = \frac{1}{Z(\thetav^t)} \exp \left(
    \sum_{(u,v) \in E^t} \theta_{uv}^{t}x_ux_v \right), \quad
  \forall t \in \mathcal{T}_n,
\end{equation}
%%%
where $Z(\thetav^t)$ is the partition function, $\thetav^t \in
\mathbb{R}^{p \choose 2}$ is the parameter vector and $G^t = (V, E^t)$
is an undirected graph representing certain conditional independence
assumptions among subsets of the $p$-dimensional random vector
$\Xv^t$. For any given time point $\tau \in [0,1]$, we are interested
in estimating the graph $G^{\tau}$ associated with
$\mathbb{P}_{\thetav^\tau}$, given the observations $\mathcal{D}_n$.

Since we are primarily interested in a situation where the total
number of observation $n$ is small compared to the dimension $p$, our
estimation task is going to be feasible only under some regularity
conditions. We impose two natural assumptions: the {\it sparsity} of
the graphs $\{G^{t}\}_{t \in \mathcal{T}_ n}$, and the {\it
  smoothness} of the parameters $\thetav^t$ as functions of
time. These assumptions are precisely stated in Section~\ref{theory}.
Intuitively, the smoothness assumption is required so that a graph
structure at the time point $\tau$ can be estimated from samples close
in time to~$\tau$. On the other hand, the sparsity assumption is
required to avoid the curse of dimensionality and to ensure that a the
graph structure can be identified from a small sample.

The model given in Eq.~\eqref{eq:model_mrf_intro} can be thought of as
a nonparametric extension of conventional MRFs, in the similar way as
the varying-coefficient models \citep{Cleveland91local,
  hastie93varying} are thought of as an extension to the linear
regression models. The difference between the model given in
Eq.~\eqref{eq:model_mrf_intro} and an MRF model is that our model
allows for parameters to change, while in MRF the parameters are
considered fixed. Allowing parameters to vary over time increases the
expressiveness of the model, and make it more suitable for
longitudinal network data.  For simplicity of presentation, in this
paper we consider time-varying MRFs with only pairwise potentials as
in Eq.~\eqref{eq:model_mrf_intro}. Note that in the case of discrete
MRFs there is no loss of generality by considering only pairwise
interactions, since any MRF with higher-order interactions can be
represented with an equivalent MRF with pairwise interactions
\citep{wainwright08graphical}.

\section{Estimation Procedure}
\label{estimation}

In this section, we review the estimation procedure of
\citet{kolar08estimating}. Given a time point $\tau \in [0,1]$ and a
sequence of observations $\mathcal{D}_n = \{ \xv^t \sim
\mathbb{P}_{\thetav^t} | t \in \mathcal{T}_n \}$ with
$\mathbb{P}_{\thetav^t}$ defined Eq.~\eqref{eq:model_mrf_intro}, the
goal is to estimate the graph structure of the Markov random field
associated with the distribution $\mathbb{P}_{\thetav^{\tau}}$.  The
parameter vector $\thetav^{\tau}$ is a ${p \choose 2}$-dimensional
vector, indexed by distinct pairs of nodes, of which an element is
non-zero if and only if the corresponding edge $(u,v) \in
E^{\tau}$. The problem of recovering the graph structure $G^{\tau}$ is
equivalent to estimating the non-zero pattern of the vector
$\thetav^{\tau}$, i.e., locations of non-zero elements of
$\thetav^{\tau}$. A stronger notion of structure estimation is that of
{\it signed edge recovery} in which an edge $(u,v) \in E^\tau$ is
recovered together with the sign of the parameter
$\sign(\theta_{uv}^\tau)$. We will show that the estimation procedure
can consistently recover signed edges.

The estimation procedure is based on the neighborhood selection
technique, where the graph structure is estimated by combining the
local estimates of neighborhoods of each node. For each vertex $u \in
V$, define the set of neighboring edges $S^{\tau}(u) := \{(u,v)\ :\
(u,v) \in E^{\tau} \}$ and the set of {\it signed neighboring edges}
$S^{\tau}_{\pm}(u) := \{ (\sign(\theta_{uv}^{\tau}), (u,v))\ :\ (u,v)
\in S^{\tau}(u) \}$. The set of signed neighboring edges
$S^{\tau}_{\pm}(u)$ can be determined from the signs of elements of
the $(p-1)$-dimensional subvector of parameters $ \thetav_{ u}^{\tau}
:= \{\theta_{uv}^{\tau}\ :\ v \in V \backslash u\}$ associated with
vertex $u$. Under the model \eqref{eq:model_mrf_intro}, the
conditional distribution of $X_u^{\tau}$ given other variables
$\Xv_{\backslash u}^{\tau} := \{ X_v^\tau \ :\ v \in V\backslash u\}$
takes the form
%%%%%
\begin{equation} \label{eq:cond_likelihood}
\mathbb{P}_{\thetav_u^{\tau}}(x_{u}^{\tau} | 
   \Xv_{\backslash u}^{\tau} = \xv_{\backslash u}^{\tau}) =
\frac{ \exp (2x_u^{\tau} \langle\thetav_{ u}^{\tau}, 
     \xv_{\backslash u}^{\tau} \rangle ) } 
     { \exp (2x_u^{\tau} \langle\thetav_{ u}^{\tau},
     \xv_{\backslash u}^{\tau} \rangle ) + 1 },
\end{equation}
%%%%
where $\langle a, b \rangle = a'b$ denotes the dot product.  Observe
that the model given in \eqref{eq:cond_likelihood} can be viewed as
expressing $X_u^{\tau}$ as the response variable in the generalized
varying-coefficient models with $\Xv_{\backslash u}^{\tau}$ playing
the role of covariates. For
simplicity, we will write $\mathbb{P}_{\thetav_u^{\tau}}(x_{u}^{\tau} |
\Xv_{\backslash u}^{\tau} = \xv_{\backslash u}^{\tau})$ as
$\mathbb{P}_{\thetav_u^{\tau}}(x_{u}^{\tau} | \xv_{\backslash
  u}^{\tau})$.

Under the model given in Eq.~\eqref{eq:cond_likelihood} the
log-likelihood, for one data-point $t \in {\cal T}_n$, can be written
in the following form:
%%%%%%%
\begin{equation}
\begin{aligned}
  \gamma(\thetav_{ u}; \xv^t)  & = \log \mathbb{P}_{\thetav_u}(x_u^t |
\xv_{\backslash u}^t) \\
& = x_u^t \langle \thetav_{ u}, \xv_{\backslash u}^t \rangle
- \log \left( 
   \exp ( \langle \thetav_{ u}, \xv_{\backslash u}^t
   \rangle) + 
   \exp ( -\langle \thetav_{ u}, \xv_{\backslash u}^t
   \rangle )
\right).
\end{aligned}
\end{equation}
%%%%%%
For an arbitrary point of interest $\tau \in [0,1]$, the estimator
$\hat \thetav_{ u}^{\tau}$ of the sign-pattern of the vector
$\thetav_{ u}^{\tau}$ is defined as the solution to the following
convex program:
%%%%%
\begin{equation} \label{eq:opt_problem} 
 \hat \thetav_{ u}^{\tau} =
  \min_{\thetav_{ u} \in \mathbb{R}^{p-1}} 
  \left\{ \ell \left(\thetav_{ u};
      \mathcal{D}_n\right) + \lambda_n \norm{\thetav_{ u}}_1 \right\}
\end{equation}
%%%%%
where $ \ell(\thetav_{ u}; \mathcal{D}_n) = - \sum_{t \in
  \mathcal{T}_n} w_t^\tau \gamma(\thetav_{ u}; \xv^t)$ is the weighted
logloss, with weights defined as
%%%
$$ 
w_t^{\tau} = \frac{ K_{h} (t-\tau) }
           {\sum_{t' \in \mathcal{T}_n} K_{h}(t'-\tau)}
$$ 
%%%
and $K_{h}(\cdot) = K(\cdot/h)$ is a symmetric nonnegative kernel. The
regularization parameter $\lambda_n \geq 0$ is specified by a user and
controls the sparsity of the solution.  The
program~\eqref{eq:opt_problem} is convex and a minimum over
$\thetav_u$ is always achieved, as the problem can be cast as a
constrained optimization problem over the ball $\norm{ \thetav_u }_1
\leq C(\lambda_n)$ and the claim follows from the Weierstrass theorem.

Let $\hat \thetav_{ u}^{\tau}$ be a minimizer of
\eqref{eq:opt_problem}. The convex program \eqref{eq:opt_problem} does
not necessarily have a unique optimum, but as we will prove shortly,
in the regime of interest any two solutions will have non-zero
elements in the same positions. Based on the vector $\hat \thetav_{
  u}^{\tau}$, we have the following estimate of the signed
neighborhood:
%%%%%
\begin{equation} \label{eq:estim_signed_neighborhood}
  \hat{ S }^{\tau}_{\pm}(u) := \left\{ (\sign(\hat
    \theta^{\tau}_{ uv}),(u,v))\ :\ v \in V \backslash u,\ \hat
      \theta^{\tau}_{ uv} \neq 0
  \right\}.  
\end{equation} 
%%%%%
The structure of graph $G^{\tau}$ is consistently estimated if every
signed neighborhood is recovered, i.e. $\hat{S}_{\pm}^\tau(u) =
S_{\pm}^\tau(u)$ for all $u \in V$. A summary of the algorithm is
given in Algorithm \ref{alg:structure_estimation}.

\begin{algorithm}[t]
\caption{Graph structure estimation}
\raggedright{
  \textbf{Input}: Dataset $\mathcal{D}_n$, time point of
  interest $\tau \in [0,1]$, penalty parameter $\lambda_n$, bandwidth
  parameter $h$ \\
\textbf{Output}: Estimate of the graph structure $\hat{G}^\tau$\\[-0.3cm]}
\begin{algorithmic}[1]
\FORALL{$u \in V$}
  \STATE Estimate $\hat \thetav_{u}$ by solving the convex
    program \eqref{eq:opt_problem} 
  \STATE Estimate the set of signed neighboring edges $\hat{ S
     }_{\pm}^\tau(u)$ using \eqref{eq:estim_signed_neighborhood}
\ENDFOR
\STATE Combine sets $\{ \hat{ S }_{\pm}^\tau(u) \}_{u \in V}$ to
  obtain $\hat G^{\tau}$.
\end{algorithmic}
\label{alg:structure_estimation}
\end{algorithm}

The convex program \eqref{eq:opt_problem}, can be solved using any
general optimization solver. One particularly fast algorithm, based on
the coordinate-wise descent method, for this type of a problem is
described in \citet{Friedman08regularization} and implemented as the R
package \emph{glmnet}. Note that the algorithm provides only an
estimate of the graph structure at time point $\tau$ and in order to
get insight into the dynamics of the graph changes, one needs to
estimate the graph structure at multiple time points. Typically, in a
real application task, one is interested in estimating $G^{\tau}$ for
all $\tau \in {\cal T}_n$.

\section{Main theoretical result}
\label{theory}

In this section, we provide conditions under which
Algorithm~\ref{alg:structure_estimation} consistently recovers the
graph structure. In particular, we show that under suitable conditions
$\mathbb{P}[\forall u\ \hat S^{\tau}_{\pm}(u) = S^{\tau}_{\pm}(u)]
\xrightarrow{n \rightarrow \infty} 1$, the property known as {\it
  sparsistency}.  We are mainly interested in the high-dimensional
case, where the dimension $p = p_n$ is comparable or even larger than
the sample size $n$. It is of great interest to understand the
performance of the estimator under this assumption, since in many real
world scenarios the dimensionality of data is large. Our analysis is
asymptotic and we consider the model dimension $p = p_n$ to grow at a
certain rate as the sample size grows. This essentially allows us to
consider more ``complicated'' models as we observe more data
points. Another quantity that will describe the complexity of the
model is the maximum node degree $s = s_n$, which is also considered
as a function of the sample size. Under the assumption that the
true-graph structure is sparse, we will require that the maximum node
degree is small, $s \ll n$. The main result describes the scaling of
the triple $(n, p_n, s_n)$ under which the estimation procedure given
in the previous section estimates the graph structure consistently.

We will need certain regularity conditions to hold in order to prove
the sparsistency result. These conditions are expressed in terms of
the Hessian of the log-likelihood function as evaluated at the true
model parameter, i.e., the Fisher information matrix. The Fisher
information matrix $\Qv_u^{\tau} \in \mathbb{R}^{(p-1)\times(p-1)}$ is
a matrix defined for each node $u \in V$ as:
%%%%
\begin{equation*}
\begin{aligned}
\Qv_u^{\tau} :&= \mathbb{E}[ \nabla^2 \log\mathbb{P}_{\thetav_u^{\tau}}
  [X_u | \Xv_{\backslash u}]] \\ 
& = \mathbb{E}[\eta(\Xv;
  \thetav_u^{\tau})\Xv_{\backslash u}\Xv_{\backslash u}'],
\end{aligned}
\end{equation*}
%%%%
where 
%%%
$$
 \eta(\xv; \thetav_u) := \frac{4 \exp (2x_u \langle
  \thetav_{ u}, \xv_{\backslash u}\rangle ) }
{( \exp (2x_u \langle \thetav_{ u},
  \xv_{\backslash u}\rangle ) + 1 )^2} 
$$
%%%
is the variance function and $\nabla^2$ denotes the operator that
computes the matrix of second derivatives. We write $\Qv^{\tau} :=
\Qv_u^{\tau}$ and assume that the following assumptions hold for each
node $u \in V$.
%%%%%
\begin{description}
\item[A1: Dependency condition] There exist constants $C_{\min}, D_{\min},
  D_{\max} > 0$ such that
%%%
$$
\Lambda_{\min}(\Qv_{SS}^\tau) \geq C_{\min}
$$
%%%
and
%%%
$$
\Lambda_{\min} \left( \Sigmav^\tau\right) \geq D_{\min},\quad
\Lambda_{\max} \left( \Sigmav^\tau\right) \leq D_{\max},
$$
%%%
where $\Sigmav^\tau = \mathbb{E}_{\thetav^{\tau}}[\Xv^{\tau}
\Xv^{\tau'}]$. Here $\Lambda_{\min}(\cdot)$ and
$\Lambda_{\max}(\cdot)$ denote the minimum and maximum eigenvalue of a
matrix.
\item[A2: Incoherence condition] There exists an incoherence parameter
  $\alpha \in (0,1]$ such that 
%%%
$$
\opnorm{\Qv_{S^cS}^\tau(\Qv_{SS}^\tau)^{-1}}{\infty} \leq 1 - \alpha,
$$
%%%
where, for a matrix $A \in \mathbb{R}^{a \times b}$, the
$\ell_{\infty}$ matrix norm is defined as $\opnorm{A}{\infty} :=
\max_{i \in \{ 1, \ldots, a\}} \sum_{j=1}^b |a_{ij}|$. Here the set
$S^c$ denotes the complement of the set $S$ in $\{1, \ldots, p\}$,
that is, $S^c = \{1, \ldots, p\}\backslash S$. 
\end{description}
%%%%%
With some abuse of notation, when defining assumptions A1 and A2, we
use the index set $S := S^\tau(u)$ to denote nodes adjacent to the
node $u$ at time~$\tau$. For example, if $s = |S|$, then $\Qv_{SS}^\tau
\in \mathbb{R}^{s \times s}$ denotes the sub-matrix of $\Qv^\tau$
indexed by $S$. 

Condition A1 assures that the relevant features are not too
correlated, while condition A2 assures that the irrelevant features do
not have to strong effect onto the relevant features. Similar
conditions are common in other literature on high-dimensional
estimation (see, e.g., \citet{meinshausen06high, ravikumar09high,
  peng09partial, guo10joint} and references therein). The difference
here is that we assume the conditions hold for the time point of
interest $\tau$ at which we want to recover the graph structure.

Next, we assume that the distribution $\mathbb{P}_{\thetav^t}$ changes
smoothly over time, which we express in the following form, for every
node $u \in V$.
%%%%
\begin{description}
\item[A3: Smoothness conditions] Let $\Sigmav^{t} =
  [\sigma_{uv}^t]$. There exists a constant $M > 0$ such that it
  upper bounds the following quantities: 
\begin{align*}
\max_{u,v \in V \times V} \sup_{t \in [0,1]} |\frac{\partial}{\partial
t} \sigma_{uv}^t| < M, \quad &
\max_{u,v \in V \times V} \sup_{t \in [0,1]} |\frac{\partial^2}{\partial
t^2} \sigma_{uv}^t| < M \\
\max_{u,v \in V \times V} \sup_{t \in [0,1]} |\frac{\partial}{\partial
t} \theta_{uv}^t| < M, \quad
&\max_{u,v \in V \times V} \sup_{t \in [0,1]} |\frac{\partial^2}{\partial
t^2}\theta_{uv}^t| < M.
\end{align*}
\end{description}
%%%%
The condition A3 captures our notion of the distribution that
changes smoothly over time. If we consider the elements of the
covariance matrix and the elements of the parameter vector as a
function of time, then these functions have bounded first and second
derivatives. From these assumptions, it is not too hard to see that
elements of the Fisher information matrix are also smooth functions of
time. 

\begin{description}
\item[A4: Kernel] The kernel $K : \mathbb{R} \mapsto \mathbb{R}$ is a
  symmetric function, supported in $[-1,1]$, and there exists a
  constant $M_K \geq 1$ which upper bounds the quantities $\max_{z \in
    \mathbb{R}} |K(z)|$ and $\max_{z \in \mathbb{R}} K(z)^2$.
\end{description}

This condition, A4, gives some regularity conditions on the kernel
used to define the weights. For example, the assumption is satisfied
by the box kernel $K(z) = \frac{1}{2}\ind\{z \in [-1,1]\}$. 

With the assumptions made above, we are ready to state the theorem
that characterizes the consistency of the method given in
Section~\ref{estimation} for recovering the unknown time-varying graph
structure. An important quantity, appearing in the statement, is the
minimum value of the parameter vector that is different from zero
%%%
$$\theta_{\min} = \min_{(u,v) \in E^{\tau}} |\theta_{uv}^{\tau}|.
$$
%%%
Intuitively, the success of the recovery should depend on how hard it
is to distinguish the true non-zero parameters from noise.

%We have the following result:
%%%%%%%%%%%%%
\begin{thm} \label{thm:main} Assume that the dependency condition
  A1 holds with $C_{\min}$, $D_{\min}$ and $D_{\max}$, that for each
  node $u \in V$, the Fisher information matrix $\Qv^{\tau}$ satisfies
  the incoherence condition A2 with parameter $\alpha$, the smoothness
  assumption A3 holds with parameter $M$, and that the kernel function
  used in Algorithm \ref{alg:structure_estimation} satisfies
  assumption A4 with parameter $M_K$. Let the regularization parameter satisfy 
%%%
\begin{equation*}
\lambda_n \geq C  \frac{ \sqrt{\log p} }{n^{1/3}}
\end{equation*}
%%%
for a constant $C > 0$ independent of $(n, p, s)$.  Furthermore,
assume that the following conditions hold:
\begin{enumerate}
\item $h = \mathcal{O}(n^{-\frac{1}{3}})$
\item $s = o(n^{1/3})$, $\frac{s^3\log p}{n^{2/3}} = o(1)$
\item $\theta_{\min} = \Omega(\frac{\sqrt{s\log p}}{n^{1/3}}).$
\end{enumerate}
Then for a fixed $\tau \in [0,1]$ the estimated graph $\hat
G^\tau(\lambda_n)$ obtained through neighborhood selection satisfies
\begin{equation*}
\mathbb{P}\left[ \hat G^\tau(\lambda_n) \neq G^{\tau} \right] =
\mathcal{O} \left( \exp \left( -C\frac{n^{2/3}}{s^3} + C'\log p \right)
\right) \rightarrow 0,
\end{equation*}
for some constants $C', C''$ independent of $(n, p, s)$.
\end{thm}
%%%%%%%%%%%%%
This theorem guarantees that the procedure in Algorithm~\ref{alg:structure_estimation}
asymptotically recovers the sequence of graphs underlying all the
nodal-state measurements in a time series, and the snapshot of the
evolving graph at any time point during measurement intervals, under
appropriate regularization parameter $\lambda_n$ as long as the
ambient dimensionality $p$ and the maximum node degree $s$ are not too
large, and minimum $\thetav$ values do not tend to zero too fast. 

{\bf Remarks:}
\begin{enumerate}
\item The bandwidth parameter $h$ is chosen so that it balances
  variance and squared bias of estimation of the elements of the
  Fisher information matrix.
\item Theorem~\ref{thm:main} states that the tuning parameter
  $\lambda$ can be set as $\lambda_n \geq Cn^{-1/3}\sqrt{\log p}$. In
  practice, one can use the Bayesian information criterion to select
  the tuning parameter $\lambda_n$ is a data dependent way, as
  explained in Section 2.4 of \cite{kolar08estimating}. We conjecture
  that this approach would lead to asymptotically consistent model
  selection, however, this claim needs to be proven.
\item Condition 2 requires that the size of the neighborhood of each
  node remains smaller than the size of the samples. However, the
  model ambient dimension $p$ is allowed to grow exponentially in
  $n$.
\item Condition 3 is crucial to be able to distinguish true elements
  in the neighborhood of a node. We require that the size of the
  minimum element of the parameter vector stays bounded away from
  zero.
\item The rate of convergence is dictated by the rate of convergence
  of the sample Fisher information matrix to the true Fisher
  information matrix, as shown in Lemma
  \ref{lem:proof:sample_fisher_information}. Using a local linear
  smoother, instead of the kernel smoother, to estimate the
  coefficients in the model \eqref{eq:cond_likelihood} one could get a
  faster rate of convergence.
\item Theorem~\ref{thm:main} provides sufficient conditions for
  reliable estimation of the sequence of graphs when the sample size
  is large enough. In order to improve small sample properties of the
  procedure, one could adapt the approach of \cite{guo2010joint} to
  the time-varying setting, to incorporate sharing between nodes. 
  \cite{guo2010joint} estimate all the local neighborhoods
  simultaneously, as opposed to estimating each neighborhood
  individually, effectively reducing the number of parameters needed
  to be inferred from data. This is especially beneficial in networks
  with prominent hubs and scale-free networks. 
\end{enumerate}

In order to obtain insight into the network dynamics one needs to
estimate the graph structure at multiple time points. A common choice
is to estimate the graph structure for every $\tau \in {\cal T}_n$ and
obtain a sequence of graph structures $\{\hat G^\tau\}_{\tau \in {\cal
    T}_n}$. We a have the following immediate consequence of
Theorem~\ref{thm:main}.

\begin{cor}
  Under the assumptions of Theorem~\ref{thm:main}, we have that 
  \begin{equation}
    \mathbb{P}\left[\forall \tau \in {\cal T}_n\ :\ 
      \hat G^\tau(\lambda_n) = G^{\tau} \right] 
      \xrightarrow{n \rightarrow \infty} 1.
  \end{equation}
\end{cor}

In the sequel, we set out to prove Theorem~\ref{thm:main}. First, we
show that the minimizer $\hat \thetav_u^\tau$ of
\eqref{eq:opt_problem} is unique under the assumptions given in
Theorem~\ref{thm:main}. Next, we show that with high probability the
estimator $\hat \thetav_u^\tau$ recovers the true neighborhood of a
node $u$. Repeating the procedure for all nodes $u \in V$ we obtain
the result stated in Theorem~\ref{thm:main}. The proof uses the
results that the empirical estimates of the Fisher information matrix
and the covariance matrix are close elementwise to their population
versions. These results are given in Appendix~\ref{appendix-a}.

\section{Proof of the main result}
\label{proof}

In this section we give the proof of Theorem~\ref{thm:main}. The proof
is given through a sequence of technical lemmas. We build on the ideas
developed in \citet{ravikumar09high}. Note that in what follows, we
use $C, C'$ and $C''$ to denote positive constants independent of
$(n,p,s)$ and their value my change from line to line.

The main idea behind the proof is to characterize the minimum obtained
in Eq.~\eqref{eq:opt_problem} and show that the correct neighborhood
of one node at an arbitrary time point can be recovered with high
probability. Next, using the union bound over the nodes of a graph, we
can conclude that the whole graph is estimated sparsistently at the
time points of interest.

We first address the problem of uniqueness of the solution to
\eqref{eq:opt_problem}. Note that because the objective in
Eq.~\eqref{eq:opt_problem} is not strictly convex, it is necessary to
show that the non-zero pattern of the parameter vector is unique,
since otherwise the problem of sparsistent graph estimation would be
meaningless. Under the conditions of Theorem \ref{thm:main} we have
that the solution is unique. This is shown in
Lemma~\ref{lem:same_non_zero} and
Lemma~\ref{lem:proof:eigen_covariance}. Lemma~\ref{lem:same_non_zero}
gives conditions under which two solutions to the problem in
Eq.~\eqref{eq:opt_problem} have the same pattern of non-zero
elements. Lemma~\ref{lem:proof:eigen_covariance} then shows, that with
probability tending to $1$, the solution is unique. Once we have shown
that the solution to the problem in Eq.~\eqref{eq:opt_problem} is
unique, we proceed to show that it recovers the correct pattern of
non-zero elements. To show that, we require the sample version of the
Fisher information matrix to satisfy certain conditions. Under the
assumptions of Theorem~\ref{thm:main},
Lemma~\ref{lem:proof:sample_fisher_information} shows that the sample
version of the Fisher information matrix satisfies the same conditions
as the true Fisher information matrix, although with worse
constants. Next we identify two events, related to the
Karush-Kuhn-Tucker optimality conditions, on which the vector $\hat
\thetav_u$ recovers the correct neighborhood the node $u$. This is
shown in Proposition~\ref{prop:sign_consistency}. Finally,
Proposition~\ref{prop:high_probability_event} shows that the event, on
which the neighborhood of the node $u$ is correctly identified, occurs
with probability tending to $1$ under the assumptions of
Theorem~\ref{thm:main}. Table~\ref{table:summary_results} provides a
summary of different parts of the proof.

{
\setlength{\tabcolsep}{12pt}
\begin{center}
\begin{table}
    \caption{
      \label{table:summary_results}
      Outline of the proof strategy.}
  \begin{tabular}{l p{7cm}}   
    \hline\hline
    Result & Description of the result\\
    \hline
    \hline
    Lemma~\ref{lem:same_non_zero}
    and Lemma~\ref{lem:proof:eigen_covariance}
    & These two lemmas establish the uniqueness of the solution to the
    optimization problem in Eq.~\eqref{eq:opt_problem}. \\
    \hline
    Lemma~\ref{lem:proof:sample_fisher_information}&
    Shows that the sample version of the
    Fisher information matrix satisfies the similar conditions to the
    population version of the Fisher information matrix.\\
    \hline
    Proposition~\ref{prop:sign_consistency} &
    Shows that on an event, related to the KKT conditions, 
    the vector $\hat
    \thetav_u$ recovers the correct neighborhood the node $u$.\\
    \hline
    Proposition~\ref{prop:high_probability_event} &
    Shows that the event in Proposition~\ref{prop:sign_consistency}
    holds with probability tending to $1$.\\
    \hline
    \hline
  \end{tabular}
\end{table}

\end{center}
}

Let us denote the set of all solution to \eqref{eq:opt_problem} as
$\Theta(\lambda_n)$. We define the objective function in
Eq.~\eqref{eq:opt_problem} by
%%%%%%%%%%%%
\begin{equation} \label{eq:proof:def_F}
F(\thetav_u) := -\sum_{t \in \mathcal{T}_n} w_t^\tau \gamma(\thetav_u; \xv^t) + 
\lambda_n\norm{\thetav_u}_1
\end{equation}
%%%%%%%%%%%%
and we say that $\thetav_u \in \mathbb{R}^{p - 1}$ satisfies the system
($\mathcal{S}$) when
%%%%
\begin{equation} \label{eq:kkt}
\forall v=1,\ldots,p-1,
\left\{
\begin{array}{cl}
\sum_{t \in \mathcal{T}_n} w_t^\tau(\nabla \gamma(\thetav_u; \xv^t))_v = 
\lambda_n \sign(\theta_{uv}) &\quad \text{if }\theta_{uv} \neq 0 \\
|\sum_{t \in \mathcal{T}_n} w_t^\tau(\nabla \gamma(\thetav_u; \xv^t))_v| \leq 
\lambda_n & \quad \text{if }\theta_{uv} = 0,
\end{array}
\right.
\end{equation}
%%%%
where
%%%%
\begin{equation} \label{eq:score}
\nabla \gamma(\thetav_u; \xv^t) = 
\xv_{\backslash u}^t \left\{ x_u^t + 1 - 
2\mathbb{P}_{\thetav_u}[x_u^t = 1 | \xv_{\backslash u}^t]
\right\}
\end{equation}
%%%%
is the score function. Eq.~\eqref{eq:kkt} is obtained by taking the
sub-gradient of $F(\thetav)$ and equating it to zero. From the
Karush-Kuhn-Tucker (KKT) conditions it follows that $\thetav_u
\in\mathbb{R}^{p-1}$ belongs to $\Theta(\lambda_n)$ if and only if
$\thetav_u$ satisfies the system~($\mathcal{S}$). The following Lemma
shows that any two solutions have the same non-zero pattern.

\begin{lem} \label{lem:same_non_zero} Consider a node $u \in V$. If
  $\bar \thetav_u \in \mathbb{R}^{p-1}$ and $\tilde{\thetav}_u \in
  \mathbb{R}^{p-1}$ both belong to $\Theta(\lambda_n)$ then $\langle
  \xv_{\backslash u}^t, \bar{\thetav}_u \rangle = \langle
  \xv_{\backslash u}^t, \tilde{\thetav}_u \rangle$, $t \in {\cal
    T}_n$. Furthermore, solutions $\bar \thetav_u$ and $\tilde{\thetav}_u$
  have non-zero elements in the same positions.
\end{lem}

We now use the result of Lemma \ref{lem:same_non_zero} to show that
with high probability the minimizer in \eqref{eq:opt_problem} is
unique. We consider the following event:
%%%%%%
$$
\Omega_{01} = \{ D_{\min} - \delta \leq 
\yv' \hat \Sigmav_{SS}^\tau \yv \leq D_{\max} + \delta \ :\ \yv \in
\mathbb{R}^{s}, \norm{\yv}_2 = 1 \}.
$$
%%%%%%
\begin{lem} \label{lem:proof:eigen_covariance}
Consider a node $u \in V$. Assume that the conditions of Lemma
\ref{lem:covariance} are satisfied. Assume also that the dependency
condition A1 holds. There are constants $C, C', C'' >
0$ depending on $M$ and $M_K$ only, such that 
%%%%%%
$$
\mathbb{P}[\Omega_{01}] \geq
1 - 4\exp(-Cnh(\frac{\delta}{s} - C'h )^2 + C''\log(s)).
$$
%%%%%%
Moreover, on the event $\Omega_{01}$, the minimizer of
\eqref{eq:opt_problem} is unique.
\end{lem}

We have shown that the estimate $\hat \thetav_u^\tau$ is unique on the
event $\Omega_{01}$, which under the conditions of
Theorem~\ref{thm:main} happens with probability converging to~1
exponentially fast. To finish the proof of Theorem~\ref{thm:main} we
need to show that the estimate $\hat \thetav_u^\tau$ has the same
non-zero pattern as the true parameter vector $\thetav_u^{\tau}$. In
order to show that we consider a few ``good'' events, which happen
with high probability and on which the estimate $\hat \thetav_u^\tau$
has the desired properties. We start by characterizing the sample
version of the Fisher information matrix, defined in
Eq.~\eqref{eq:sample_fisher}. Consider the following events:
%%%%%
$$ \Omega_{02} := \{ C_{\min} - \delta \leq \yv' \hat \Qv_{SS}^\tau \yv
\ :\ \yv \in \mathbb{R}^{s}, \norm{\yv}_2 = 1 \}
$$
%%%%%
and
%%%%%
$$
\Omega_{03} := \{ \opnorm{\hat \Qv_{S^cS}^\tau(\hat
  \Qv_{SS}^\tau)^{-1}}{\infty} \leq 1 - \frac{\alpha}{2} \}.
$$
%%%%%
\begin{lem} \label{lem:proof:sample_fisher_information} Assume that
  the conditions of Lemma \ref{lem:covariance} are satisfied. Assume
  also that the dependency condition A1 holds and the incoherence
  condition A2 holds with the incoherence parameter $\alpha$. There
  are constants $C, C', C'' > 0$ depending on $M$, $M_K$ and $\alpha$
  only, such that
%%%%
$$
\mathbb{P}[\Omega_{02}] \geq 1 - 2\exp(-C\frac{nh\delta^2}{s^2} +
C'\log(s))
$$
%%%%
and
%%%%
$$
\mathbb{P}[\Omega_{03}] \geq  1 - \exp(-C\frac{nh}{s^3} +
C''\log(p)). 
$$
%%%%
\end{lem}

Lemma \ref{lem:proof:sample_fisher_information} guarantees that the
sample Fisher information matrix satisfies ``good'' properties with
high probability, under the appropriate scaling of quantities $n, p,
s$ and $h$. 

We are now ready to analyze the optimum to the convex program
\eqref{eq:opt_problem}. To that end we apply the mean-value theorem
coordinate-wise to the gradient of the weighted logloss $\sum_{t \in
  \mathcal{T}_n} w_t^\tau \nabla \gamma(\thetav_u; \xv^t)$ and obtain
%%%%
\begin{equation} \label{eq:proof:mean_value}
\sum_{t \in \mathcal{T}_n} w_t^\tau(
  \nabla \gamma(\hat \thetav_u^\tau; \xv^t) - 
  \nabla \gamma(\thetav_u^{\tau}; \xv^t) ) = 
[\sum_{t \in \mathcal{T}_n} w_t^\tau \nabla^2 
   \gamma(\thetav_u^{\tau};\xv^t)]
(\hat \thetav_u^\tau - \thetav_u^{ \tau}) + \Deltav^\tau,
\end{equation}
%%%%
where $\Deltav^\tau \in \mathbb{R}^{p -1}$ is the remainder term of the
form
%%%%
\begin{equation}
\Delta_v^\tau\ =\ [\sum_{t \in \mathcal{T}_n} w_t^\tau(
   \nabla^2 \gamma(\bar{\thetav}_u^{(v)}; \xv^t) - 
   \nabla^2 \gamma(\thetav_u^{\tau}; \xv^t))]_v'
 (\hat \thetav_u^\tau - \thetav_u^{\tau})
\end{equation}
%%%%
and $\bar{\thetav}_u^{(v)}$ is a point on the line between
$\thetav_u^{\tau}$ and $\hat \thetav_u^\tau$, and $[\cdot]_v'$ denoting
the $v$-th row of the matrix. Recall that $\hat \Qv^\tau = \sum_{t \in
  \mathcal{T}_n} w_t^\tau \nabla^2 \gamma(\thetav_u^{\tau}; \xv^t)$.
Using the expansion \eqref{eq:proof:mean_value}, we write the KKT
conditions given in Eq.~\eqref{eq:kkt} in the following form, $\forall
v=1,\ldots,p-1$,
%%%%
\begin{equation} \label{eq:proof:kkt}
\left\{
\begin{array}{cl}
\hat \Qv_v^\tau(\thetav_u - \thetav_u^{\tau}) +
\sum_{t \in \mathcal{T}_n} w_t^\tau(\nabla \gamma(\thetav_u^{\tau}; \xv^t) )_v
 + \Delta_v^\tau= 
\lambda_n \sign(\theta_{uv}) 
&\quad \text{if }\theta_{uv} \neq 0 \\
|\hat \Qv_v^\tau(\thetav_u - \thetav_u^{\tau}) + 
\sum_{t \in \mathcal{T}_n} w_t^\tau(\nabla \gamma(\thetav_u^{\tau}; \xv^t) )_v
+ \Delta_v^\tau | \leq 
\lambda_n & \quad \text{if }\theta_{uv} = 0.
\end{array}
\right.
\end{equation}
%%%%
We consider the following events
%%%%
$$
\Omega_0 = \Omega_{01} \cap \Omega_{02} \cap \Omega_{03},
$$
%%%%
$$
\Omega_1 = \{ \forall v \in S\ :\  
|\lambda_n ((\hat{\Qv}_{SS}^{\tau})^{-1} \sign(\thetav_S^{\tau}))_v - 
((\hat{\Qv}_{SS}^\tau)^{-1}\Wv_S^\tau)_v| 
< |\theta_{uv}^{\tau}|\}
$$
%%%%
and
%%%%
$$
\Omega_2 = \{ \forall v \in S^c\ :\ |(\Wv_{S^c}^\tau
-\hat{\Qv}_{S^cS}^\tau(\hat{\Qv}_{SS}^\tau)^{-1}\Wv_S^\tau)_v| <
\frac{\alpha}{2}\lambda_n \}
$$
%%%%
where 
%%%
$$
\Wv^\tau = \sum_{t \in \mathcal{T}_n} w_t^\tau \nabla
\gamma(\thetav_u^{\tau}; \xv^t) +\Deltav^\tau.
$$
%%%
We will work on the event $\Omega_0$ on which the minimum eigenvalue
of $\hat{\Qv}_{SS}^\tau$ is strictly positive and, so,
$\hat{\Qv}_{SS}^\tau$ is regular and $\Omega_0 \cap \Omega_1$ and
$\Omega_0 \cap \Omega_2$ are well defined.
%%%%
\begin{prop} \label{prop:sign_consistency} Assume that the conditions
  of Lemma \ref{lem:proof:sample_fisher_information} are satisfied. The
  event
%%%%
$$
\{ \forall \hat \thetav_u^\tau \in
\mathbb{R}^{p-1} \text{ solution of }({\cal S}), \text{ we have }
\sign(\hat \thetav_u^\tau) = \sign(\thetav_u^{\tau})\} \cap \Omega_0
$$ 
%%%%
contains event $\Omega_0 \cap \Omega_1 \cap \Omega_2$.
\end{prop}
%%%%

\begin{proof}
  We consider the following linear functional
%%%%%%%%%%%%%%%%%%%%
\begin{equation*}
  G : \left\{
    \begin{array}{ccl}
      \mathbb{R}^{s} & \rightarrow & \mathbb{R}^{s}\\
      \thetav & \mapsto & \thetav - \thetav_S^{\tau} +
      (\hat \Qv_{SS}^\tau)^{-1}\Wv_S^\tau -
      \lambda_n (\hat \Qv_{SS}^\tau)^{-1}\sign(\thetav_S^{\tau}). \\
    \end{array}
  \right.
\end{equation*}
%%%%%%%%%%%%%%%%%%%%
For any two vectors $\yv = (y_1, \ldots, y_{s})' \in \mathbb{R}^{s}$
and $\rv = (r_1, \ldots, r_{s})' \in \mathbb{R}_+^{s}$, define the
following set centered at $\yv$ as
%%%
$$
\mathcal{B}(\yv,\rv)= \prod_{i=1}^{s} (y_i - r_i, y_i + r_i).
$$
%%%
Now, we have
%%%%%%%%%%%%%%%%%%%%
$$
G \left(
   \mathcal{B}(\thetav_S^{\tau},|\thetav_S^{\tau}|) 
\right) = 
\mathcal{B} \left(
(\hat \Qv_{SS}^\tau)^{-1}\Wv_S^\tau -
\lambda_n (\hat \Qv_{SS}^\tau)^{-1}\sign(\thetav_S^{\tau}),
|\thetav_S^{\tau}| \right).
$$
%%%%%%%%%%%%%%%%%%%%
On the event $\Omega_0 \cap \Omega_1$, 
%%%
$$
0 \in \mathcal{B} \left( (\hat \Qv_{SS}^\tau)^{-1}\Wv_S^\tau -
  \lambda_n (\hat \Qv_{SS}^\tau)^{-1} \sign(\thetav_S^{\tau}),
  |\thetav_S^{\tau}| \right),
$$
%%%
which implies that there exists a vector $\bar{\thetav}_{S}^\tau \in
\mathcal{B}(\thetav_S^{\tau},|\thetav_S^{\tau}|)$ such that
$G(\bar{\thetav}_{S}^\tau) = 0$. For $\bar{\thetav}_{S}^\tau$ it holds
that $\bar{\thetav}_{S}^\tau = \thetav_S^{\tau} + \lambda_n (\hat
\Qv_{SS}^\tau)^{-1} \sign(\thetav_S^{\tau}) - (\hat \Qv_{SS}^\tau)^{-1}
\Wv_S^\tau$ and $|\bar{\thetav}_{S}^\tau - \thetav_S^{\tau}| <
|\thetav_S^\tau|$. Thus, the vector $\bar{\thetav}_{S}^\tau$ satisfies
%%%%%%
$$\sign(\bar{\thetav}_{S}^\tau) = \sign(\thetav_S^{\tau})
$$
%%%%%%
 and
%%%%%%
\begin{equation} \label{eq:proof:kkt_equiv_1}
 \hat \Qv_{SS}(\bar{\thetav}_S^\tau - \thetav_S^{\tau}) + \Wv_S^\tau = 
\lambda_n \sign(\bar{\thetav}_S^\tau).
\end{equation}
%%%%%%

Next, we consider the vector $\bar{\thetav}^\tau = \left(
\begin{array}{c}
\bar{\thetav}_{S}^\tau \\
\bar{\thetav}_{S^c}^\tau
\end{array}
\right) $ 
%%%
where $\bar{\thetav}_{S^c}^\tau$ is the null vector of $\mathbb{R}^{p-1-s}$.
On event $\Omega_0$, from Lemma
\ref{lem:proof:sample_fisher_information} we know that $\opnorm{\hat
  \Qv_{S^cS}^\tau(\hat \Qv_{SS}^\tau)^{-1}}{\infty} \leq 1 -
\frac{\alpha}{2}$. Now, on the event $\Omega_0 \cap \Omega_2$ it holds
%%%%%%
\begin{equation} \label{eq:proof:kkt_equiv_2}
\begin{aligned}
  & \norm{\hat \Qv_{S^c S}^\tau(\bar \thetav_S^\tau - \thetav_S^{\tau})
    + \Wv_{S^c}^\tau}_{\infty}
  = \\
  & \quad \norm{-\hat \Qv_{S^cS}^\tau (\hat{\Qv}_{SS}^\tau)^{-1}
    \Wv_S^\tau + \Wv_{S^c}^\tau + \lambda_n \hat \Qv_{S^cS}^\tau
    (\hat{\Qv}_{SS}^\tau)^{-1}\sign(\bar{\thetav}_S^\tau) }_{\infty} < \lambda_n.
\end{aligned}
\end{equation}
%%%%%%
Note that for $\bar{\thetav}^\tau$, equations
\eqref{eq:proof:kkt_equiv_1} and \eqref{eq:proof:kkt_equiv_2} are
equivalent to saying that $\bar{\thetav}^\tau$ satisfies conditions
\eqref{eq:proof:kkt} or \eqref{eq:kkt}, i.e., saying that
$\bar{\thetav}^\tau$ satisfies the KKT conditions. Since
$\sign(\bar{\thetav}_{S}^\tau) = \sign(\thetav_S^{\tau})$, we have
$\sign(\bar{\thetav}^\tau) = \sign(\thetav_u^{\tau})$. Furthermore,
because of the uniqueness of the solution to \eqref{eq:opt_problem} on
the event $\Omega_0$ , we conclude that $\hat{\thetav}_u^\tau = \bar
\thetav^\tau$.
\end{proof}

Proposition \ref{prop:sign_consistency} implies Theorem \ref{thm:main}
if we manage to show that the event $\Omega_0 \cap \Omega_{1} \cap
\Omega_{2}$ occurs with high probability under the assumptions stated
in Theorem \ref{thm:main}. Proposition
\ref{prop:high_probability_event} characterizes the probability of
that event, which concludes the proof of Theorem \ref{thm:main}.

\begin{prop} \label{prop:high_probability_event} Assume that the
  conditions of Theorem~\ref{thm:main} are satisfied. Then there are
  constants $C, C' > 0$ depending on $M$, $M_K$, $D_{\max}$,
  $C_{\min}$ and $\alpha$ only, such that the following holds:
\begin{equation}
  \mathbb{P}[\Omega_0 \cap \Omega_1 \cap \Omega_2] \geq 
  1 - 2\exp(-Cnh(\lambda_n - sh)^2 + \log(p)).
\end{equation}
\end{prop}

\begin{proof} 

  We start the proof of the proposition by giving a technical lemma,
  which characterizes the distance between vectors $\hat
  \thetav_u^\tau = \bar \thetav^\tau$ and $\thetav_u^\tau$ under the
  assumptions of Theorem~\ref{thm:main}, where $\bar \thetav^\tau$ is
  constructed in the proof of
  Proposition~\ref{prop:sign_consistency}. The following lemma gives a
  bound on the distance between the vectors $\hat \thetav_S^\tau$ and
  $\thetav_S^\tau$, which we use in the proof of the proposition. The
  proof of the lemma is given in Appendix. 

%%%%%
  \begin{lem} \label{lem:converge_theta} Assume that the conditions of
    Theorem~\ref{thm:main} are satisfied. There are constants $C, C' >
    0$ depending on $M, M_K, D_{\max}, C_{\min}$ and $\alpha$ only,
    such that
    \begin{equation} \label{eq:proof:primal_bound} 
      \norm{\hat \thetav_S^\tau - \thetav_S^\tau}_2 
         \leq C \frac{\sqrt{s \log p}}{n^{1/3}}
    \end{equation}
    with probability at least $1 - \exp(-C'\log p)$.
  \end{lem}
%%%%%

Using Lemma \ref{lem:converge_theta} we can prove Proposition
\ref{prop:high_probability_event}. We start by studying the
probability of the event $\Omega_2$. We have
\begin{equation*}
  \Omega_2^C \subset \cup_{v \in S^c} \{ 
  \Wv_v + (\hat{\Qv}_{S^cS}^\tau(\hat{\Qv}_{SS}^\tau)^{-1}
  \Wv_S^\tau)_v \geq \frac{\alpha}{2}\lambda_n
  \}.
\end{equation*}
Recall that $\Wv^\tau = \sum_{t \in \mathcal{T}_n} w_t^\tau \nabla
\gamma(\thetav_u^{\tau}; \xv^t) +\Deltav^\tau$. Let us define the event
%%%
$$
\Omega_3 = \{ \max_{1 \leq v \leq p-1} |\ev_v' \sum_{t \in
  \mathcal{T}_n} w_t^\tau \nabla \gamma(\thetav^{ \tau}_u; \xv^t)| <
\frac{\alpha\lambda_n}{4(2-\alpha)}\},
$$
%%%
where $\ev_v \in \mathbb{R}^{p-1}$ is a unit vector with one at the
position $v$ and zeros elsewhere.  From the proof of
Lemma~\ref{lem:converge_theta} available in the appendix we have that
$\mathbb{P}[\Omega_3] \geq 1 - 2\exp(-C \log(p))$ and on that event
the bound given in Eq.~\eqref{eq:proof:primal_bound} holds.

On the event $\Omega_3$, we bound the remainder term
$\Deltav^\tau$. Let $g : \mathbb{R} \mapsto \mathbb{R}$ be defined as
$g(z) = \frac{4\exp(2z)}{(1+\exp(2z))^2}$. Then $\eta(\xv;\thetav_u) =
g(x_u \langle \thetav_u, \xv_{\backslash u} \rangle)$. For $v \in \{1,
\ldots, p-1\}$, using the mean value theorem it follows that
%%%%%
\begin{equation*}
\begin{aligned}
\Delta_v
& \ =\ [\sum_{t \in \mathcal{T}_n} w_t^\tau(
   \nabla^2 \gamma(\bar{\thetav}_u^{(v)}; \xv^t) - 
   \nabla^2 \gamma(\thetav_u^{\tau}; \xv^t))]_v'
  (\hat \thetav_u^\tau - \thetav_u^{\tau})\\
&\ =\ \sum_{t \in \mathcal{T}_n} w_t^\tau [ 
    \eta ( \xv^t; \bar{\thetav}_u^{(v)} ) - 
    \eta ( \xv^t; \thetav_u^{\tau} ) ] 
    [ \xv_{\backslash u}^t \xv_{\backslash u}^{t'}]_v'
  [ \hat \thetav_u^\tau - \thetav_u^{\tau} ] \\
& \ =\ \sum_{t \in \mathcal{T}_n} w_t^\tau 
  g'( \xv_u^t \langle \bar{\bar{\thetav}}_u^{(v)},
      \xv_{\backslash u}^t \rangle ) 
  [ x_u^t \xv_{\backslash u}^t ]'
  [ \bar \thetav_u^{(v)} - \thetav_u^{\tau} ] 
  [x_v^t \xv_{\backslash u}^{t'}] 
  [ \hat \thetav_u^{\tau} - \thetav_u^{\tau} ]\\
&\ =\ \sum_{t \in \mathcal{T}_n} w_t^\tau
  \{ g'( x_u^t \langle \bar{\bar{\thetav}}_u^{(v)}, \xv_{\backslash u}^t
  \rangle ) x_u^tx_v^t \}
  \{ [ \bar \thetav_u^{(v)} - \thetav_u^{\tau} ]' 
   \xv_{\backslash  u}^{t} \xv_{\backslash u}^{t'} 
  [ \hat \thetav_u^\tau - \thetav_u^{\tau}] \},
\end{aligned}
\end{equation*}
%%%
where $\bar{\bar{\thetav}}_u^{(v)}$ is another point on the line
joining $\hat \thetav_u^\tau$ and $\thetav_u^\tau$. A simple
calculation shows that $| g'( x_u^t \langle
\bar{\bar{\thetav}}_u^{(v)}, \xv_{\backslash u}^t \rangle ) x_u^tx_v^t
| \leq 1$, for all $t \in \mathcal{T}_n$, so we have
\begin{equation} \label{eq:proof:bound_remainder}
\begin{aligned}
|\Delta_v|&\ \leq\ [ \bar \thetav_u^{(v)} - \thetav_u^{\tau} ]'
\{ \sum_{t \in \mathcal{T}_n} w_t^\tau \xv_{\backslash u}^{t} 
   \xv_{\backslash u}^{t'} \} [ \hat \thetav_u^\tau - \thetav_u^{\tau} ]\\
&\ \leq\ [ \hat \thetav_u^\tau - \thetav_u^{\tau} ]'
  \{ \sum_{t \in \mathcal{T}_n} w_t^\tau \xv_{\backslash u}^{t} 
  \xv_{\backslash u}^{t'} \} [ \hat \thetav_u^\tau - \thetav_u^{\tau} ] \\
&\ =\ [ \hat \thetav_S^\tau - \thetav_S^{\tau} ]'
\{ \sum_{t \in \mathcal{T}_n} w_t^\tau \xv_{S}^{t} 
  \xv_{S}^{t'} \} [ \hat \thetav_S^\tau - \thetav_S^{\tau} ] \\
&\ \leq\ D_{\max} \norm{\hat \thetav_S^\tau - \thetav_S^{\tau}}_2^2.
\end{aligned}
\end{equation}
Combining the equations \eqref{eq:proof:bound_remainder} and
\eqref{eq:proof:primal_bound}, we have that on the event $\Omega_3$
\begin{equation*}
  \max_{1 \leq v \leq p-1} |\Delta_v| \leq  C \lambda_n^2 s <
  \frac{\lambda_n \alpha}{4(2-\alpha)}
\end{equation*}
where $C$ is a constant depending on $D_{\max}$ and $C_{\min}$
only. 

On the event $\Omega_0 \cap \Omega_3$, we have
\begin{equation*}
    W_v^\tau + (\hat{\Qv}_{S^cS}^\tau(\hat{\Qv}_{SS}^\tau)^{-1}\Wv_S^\tau)_v
     < \frac{\alpha\lambda_n}{2(2-\alpha)} +
        (1-\alpha)\frac{\alpha\lambda_n}{2(2-\alpha)} 
     \leq \frac{\alpha\lambda_n}{2}
\end{equation*}
and we can conclude that $\mathbb{P}[\Omega_2] \geq 1 -
2\exp(-C\log(p))$ for some constant $C$ depending on $M, M_K,
C_{\min}, D_{\max}$ and $\alpha$ only.

Next, we study the probability of the event $\Omega_1$. We have
\begin{equation}
  \Omega_1^C \subset \cup_{v \in S} \{
  \lambda_n ((\hat{\Qv}_{SS}^\tau)^{-1} \sign(\thetav_S^{\tau}))_v +
  ((\hat{\Qv}_{SS}^\tau)^{-1}W_S^\tau)_v \geq \theta_{uv}^{\tau} \}.
\end{equation}
Again, we will consider the event $\Omega_3$. On the event $\Omega_0
\cap \Omega_3$ we have that 
\begin{equation}
  \lambda_n ((\hat{\Qv}_{SS}^\tau)^{-1} \sign(\thetav_S^{\tau}))_v +
  ((\hat{\Qv}_{SS}^\tau)^{-1}\Wv_S^\tau)_v \leq 
  \frac{\lambda_n\sqrt{s}}{C_{\min}} +
  \frac{\lambda_n}{2C_{\min}} \leq C\lambda_n\sqrt{s},
\end{equation}
for some constant $C$. When $\theta_{\min} > C\lambda_n\sqrt{s}$, we
have that $\mathbb{P}[\Omega_1] \geq 1 - 2\exp(-C\log(p))$ for some
constant $C$ that depends on $M, M_K, C_{\min}, D_{\max}$ and $\alpha$
only.
\end{proof}

In summary, under the assumptions of Theorem~\ref{thm:main}, the
probability of event $\Omega_0 \cap \Omega_1 \cap \Omega_2$ converges
to one exponentially fast. On this event, we have shown that the
estimator $\hat \thetav_u^\tau$ is the unique minimizer
of~\eqref{eq:opt_problem} and that it consistently estimates the
signed non-zero pattern of the true parameter vector $\thetav_u^\tau$,
i.e., it consistently estimates the neighborhood of a node
$u$. Applying the union bound over all nodes $u \in V$, we can
conclude that our estimation procedure explained in
Section~\ref{estimation} consistently estimates the graph structure at
a time point $\tau$.

\section{Numerical simulation}
\label{simulation}

In this section, we demonstrate numerical performance of
Algorithm~\ref{alg:structure_estimation}. A detailed comparison with
other estimation procedures and an application to biological data has
been reported in \citet{kolar08estimating}. We will use three
different types of graph structures: a chain, a nearest-neighbor and a
random graph. Each graph has $p=50$ nodes and the maximum node degree
is bounded by $s=4$. These graphs are detailed below:

\noindent{\bf Example 1: Chain graph.} First a random permutation
$\pi$ of $\{1, \ldots, p\}$ is chosen. Then a graph structure is
created by connecting consecutive nodes in the permutation, that is,
$(\pi(1), \pi(2)), \ldots, (\pi(p-1), \pi(p)) \in E$. 

\noindent{\bf Example 2: Nearest neighbor graph.} A nearest
neighbor graph if generated following the procedure outlined in
\cite{li06gradient}. For each node, we draw a point uniformly at
random on a unit square and compute the pairwise distances between
nodes. Each node is then connected to 4 closest neighbors. Since some
of nodes will have more than 4 adjacent edges, we remove randomly
edges from nodes that have degree larger than 4 until the maximum
degree of a node in a graph is 4.

\noindent{\bf Example 3: Random graph.} To generate a random graph
with $e=45$ edges, we add each edges one at a time, between random pairs
of nodes that have the node degree less than 4.

We use the above described procedure to create the first random graph
$\tilde G^0$. Next, we randomly add 10 edges and remove 10 edges from
$\tilde G^0$, taking care that the maximum node degree is still $4$,
to obtain $\tilde G^1$. Repeat the process of adding and removing
edges from $\tilde G^1$ to obtain $\tilde G^2, \ldots, \tilde G^5$. We
refer to these 6 graphs as the anchor graphs. We will randomly
generate the prototype parameter vectors $\tilde \thetav^0, \ldots,
\tilde \thetav^5$, corresponding to the anchor graphs, and then
interpolate $200$ points between them to obtain the parameters
$\{\thetav^t\}_{t\in\mathcal{T}_n}$, which gives us $n = 1000$. We
generate a prototype parameter vector $\tilde \thetav^i$ for each
anchor graph $\tilde G^i$, $i \in \{0, \ldots, 5\}$, by sampling
non-zero elements of the vector independently from ${\sf Unif}([-1,
0.5] \cup [0.5, 1])$. Now, for each $t \in \mathcal{T}_n$ we
generate $10$ {\it i.i.d.} samples using Gibbs sampling from the
distribution $\mathbb{P}_{\thetav^t}$. Specifically, we discard
samples from the first $10^4$ iterations and collect samples every
$100$ iterations.

We estimate $\hat G^t$ for each $t \in {\cal T}_n$ using $k \in \{1,
\ldots, 10\}$ samples at each time point. The results are expressed in
terms of the precision~$({\sf Pre})$ and the recall~$({\sf Rec})$ and
$F1$ score, which is the harmonic mean of precision and recall, i.e.,
$F1 := 2 * {\sf Pre} * {\sf Rec} / ({\sf Pre} + {\sf Rec})$. Let $\hat
E^t$ denote the estimated edge set of $\hat G^t$, then the precision
is calculated as ${\sf Pre} := 1/n \sum_{t \in {\cal T}_n} |\hat E^t
\cap E^t|/|\hat E^t|$ and the recall as ${\sf Rec} := 1/n \sum_{t \in
  {\cal T}_n} |\hat E^t \cap E^t|/|E^t|$. Furthermore, we report
results averaged over $100$ independent runs. The tuning parameters
are selected by maximizing the BIC score over a grid of regularization
parameters as described in
\citet{kolar08estimating}. Table~\ref{table} contains a summary of
simulation results.

\begin{table}
    \caption{
      \label{table}
      Summary of simulation results. The number of nodes $p=50$ and
      the number of discrete time points $n=1000$.}
  \begin{tabular}{c l c c c c c c c c c c}   
    \hline\hline
    &&\multicolumn{10}{c}{Number of independent samples}\\
    && 1 & 2 & 3 & 4 & 5 & 6 & 7 & 8 & 9 & 10 \\
    \hline
    \multirow{3}{*}{Precision} & 
    Chain & 
    0.75 & 0.95 & 0.96 & 0.96 & 0.97 & 0.98 & 0.99 & 0.99 & 0.99 & 0.99\\
    &NN & 
    0.84 & 0.98 & 0.97 & 0.96 & 0.98 & 0.98 & 0.98 & 0.98 & 0.97 & 0.98\\
    &Random & 
    0.55 & 0.57 & 0.65 & 0.71 & 0.75 & 0.79 & 0.83 & 0.84 & 0.85 & 0.85\\
    \\

    \multirow{3}{*}{Recall} & 
    Chain & 
    0.59 & 0.65 & 0.69 & 0.72 & 0.73 & 0.73 & 0.73 & 0.73 & 0.73 & 0.73\\
    &NN & 
    0.48 & 0.57 & 0.61 & 0.63 & 0.63 & 0.64 & 0.64 & 0.64 & 0.65 & 0.65\\
    &Random & 
    0.50 & 0.52 & 0.55 & 0.56 & 0.56 & 0.58 & 0.60 & 0.60 & 0.63 & 0.66\\
    \\

    \multirow{3}{*}{F1 score} & 
    Chain & 
    0.66 & 0.76 & 0.80 & 0.82 & 0.83 & 0.84 & 0.84 & 0.84 & 0.85 & 0.84\\
    &NN &
    0.61 & 0.72 & 0.74 & 0.76 & 0.77 & 0.77 & 0.77 & 0.77 & 0.77 & 0.78\\
    &Random & 
    0.52 & 0.54 & 0.60 & 0.63 & 0.64 & 0.67 & 0.70 & 0.70 & 0.72 & 0.74\\
    \hline
  \end{tabular}
\end{table}

\begin{figure}[t]
  \centering
  \includegraphics[width=\columnwidth]{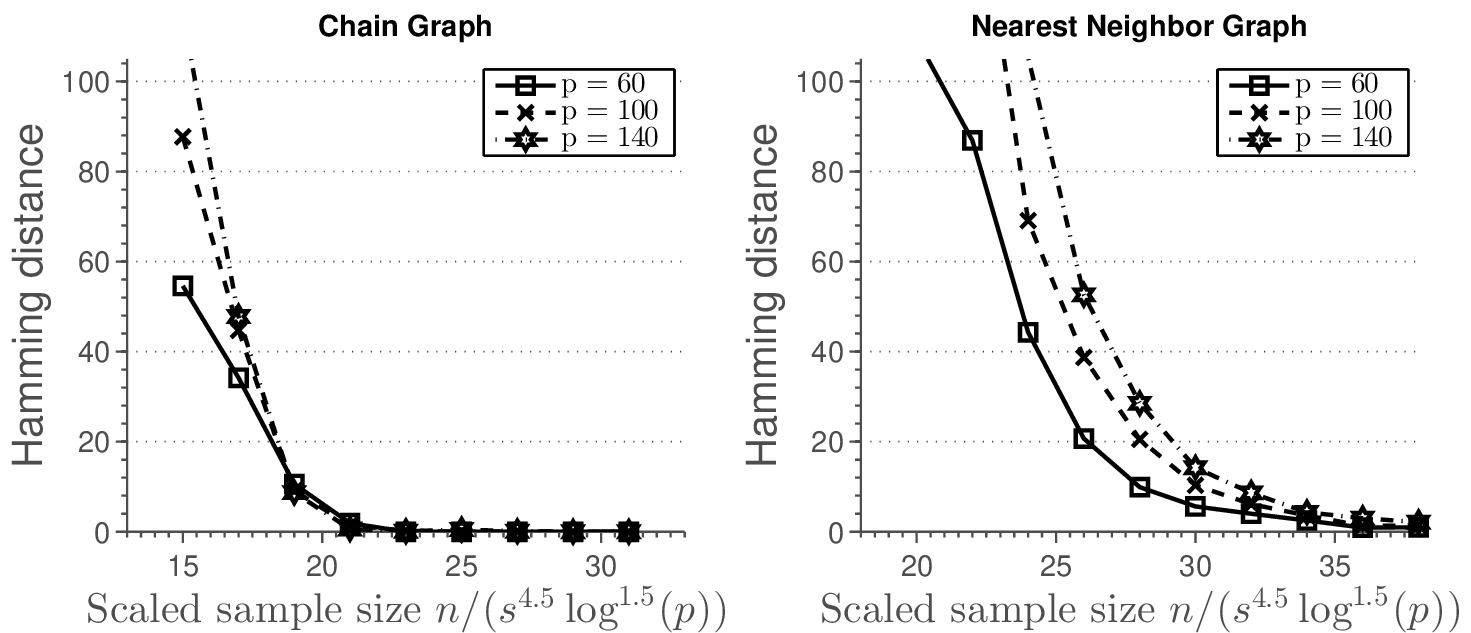}
  \caption{Average hamming distance plotted against the rescaled
    sample size. Each column represents one simulation
    setting. Results are averaged over 100 independent runs. }
  \label{fig:simulation}
\end{figure}

As suggested by the reviewer, we perform an additional simulation that
illustrates that the conditions of Theorem~\ref{thm:main} can be
satisfied. We will use the random chain graph and the nearest neighbor
graph for two simulation settings.  In each setting, we generate two
anchor graphs with $p$ nodes and create two prototype parameter
vectors, as described above. Then we interpolate these two parameters
over $n$ points. Theorem~\ref{thm:main} predicts the scaling for the
sample size $n$, as a function of other parameters, required to
successfully recover the graph at a time point $\tau$. Therefore, if
our theory correctly predicts the behavior of the estimation procedure
and we plot the hamming distance between the true and recovered graph
structure against appropriately rescaled sample size, we expect the
curves to reach zero distance for different problem sizes at a same
point. The bandwidth parameter $h$ is set as $h = 4.8n^{-1/3}$ and the
penalty parameter $\lambda_n$ as $\lambda_n = 2\sqrt{n^{-2/3}\log(p)}$
as suggested by the theory. Figure~\ref{fig:simulation} shows the
hamming distance against the scaled sample size
$n/(s^{4.5}\log^{1.5}(p))$.  Each point is averaged over 100
independent runs.

\section{Conclusion}

In the paper, we focus on sparsistent estimation of the time-varying
high-dimensional graph structure in Markov Random Fields from a small
size sample. An interesting open direction is estimation of the graph
structure from a general time-series, where observations are
dependent. In our opinion, the graph structure that changes with time
creates the biggest technical difficulties. Incorporating dependent
observations would be an easier problem to address, however, the one
of great practical importance, since samples in the real data sets are
likely to be dependent. Another open direction is to establish
necessary conditions, to complement sufficient conditions established
here, under which it is possible to estimate a time-varying graph
structure. Another research direction may be to use non-convex
penalties introduced by \cite{fan01variable} in place of the $\ell_1$
penalty. The idea would be to relax the condition imposed in the
assumption A2, since it is well known that the SCAD penalties improve
performance when the variables are correlated.

\section*{Acknowledgment}

We would like to thank Larry Wasserman for many useful discussions and
suggestions. The research reported here was supported in part by Grant
ONR N000140910758, NSF DBI-0640543, NSF DBI-0546594, NSF IIS- 0713379,
an Alfred P. Sloan Research Fellowship to EPX and a graduate
fellowship from Facebook to MK.

\appendix
\section{Large deviation inequalities}
\label{appendix-a}

In this section we characterize the deviation of elements of the
sample Fisher information matrix $\hat \Qv^\tau := \hat \Qv_u^\tau$ at
time point $\tau$, defined as
%%%
\begin{equation}
  \label{eq:sample_fisher}
  \hat \Qv^\tau = \sum_t w_t^\tau \eta(\xv^t; \thetav_u^{ \tau})\xv_{\backslash
    u}^t\xv_{\backslash u}^{t'},
\end{equation}
%%%
and the sample covariance matrix $\hat \Sigmav^\tau$ from their
population versions $\Qv^{\tau}$ and $\Sigmav^{\tau}$. These results
are crucial for the proof of the main theorem, where the consistency
result depends on the bounds on the difference $\hat \Qv^\tau -
\Qv^{\tau}$ and $\hat \Sigmav^\tau - \Sigmav^{\tau}$. In the
following, we use $C, C'$ and $C''$ as generic positive constants
independent of $(n, p, s)$.

\subsection{Sample Fisher information matrix}

To bound the deviation between elements of $\hat \Qv^\tau = [\hat
  q_{vv'}^\tau]$ and $\Qv^{\tau} = [q_{vv'}^\tau]$, $v, v' \in
V\backslash u$, we will use the following decomposition:
%%%%%%
\begin{equation} \label{eq:three_terms}
  \begin{aligned}
    |\hat q_{vv'}^\tau - q_{vv'}^\tau|\ \leq &\ 
      | \sum_{t \in \mathcal{T}_n} w_t^\tau \eta(\xv^t;\thetav_u^{\tau}) 
             x_v^tx_{v'}^t -
        \sum_{t \in \mathcal{T}_n} w_t^\tau \eta(\xv^t;\thetav_u^{t}) 
             x_v^tx_{v'}^t | \\
    + &\ 
      | \sum_{t \in \mathcal{T}_n} w_t^\tau\eta(\xv^t;\thetav_u^{t})
          x_v^tx_{v'}^t - 
        \mathbb{E}[\sum_{t \in \mathcal{T}_n}
          w_t^\tau\eta(\xv^t;\thetav_u^{t}) x_v^tx_{v'}^t]| \\
    + &\ 
      | \mathbb{E}[\sum_{t \in \mathcal{T}_n}
         w_t^\tau\eta(\xv^t;\thetav_u^{t}) x_v^tx_{v'}^t] 
         - q_{vv'}^\tau|.
  \end{aligned}
\end{equation}
%%%%%
The following lemma gives us bounds on the terms in
Eq.~\eqref{eq:three_terms}.
%%%%
\begin{lem} \label{lem:large_dev:three_terms}
Assume that the smoothness condition A3 is satisfied and that the
kernel function $K(\cdot)$ satisfies A4. Furthermore, assume 
%%%%
\begin{equation*}
  \max_{t \in [0,1]} |\{v \in \{1, \ldots, p\}\ :\ \theta_{uv}^{t} \neq 0\}| < s,
\end{equation*}
%%%%
i.e., the number of non-zero elements of the parameter vector is
bounded by $s$.  There exist constants $C, C', C'' > 0$, depending on
$M$ and $M_K$ only, which are the constants quantifying assumption A3
and A4, respectively , such that for any $\tau \in [0,1]$, we have
%%%%%
\begin{align}
\max_{v,v'}\ | \hat q_{vv'}^\tau - \sum_{t \in \mathcal{T}_n} 
  w_t^\tau \eta(\xv^t;\thetav_u^{t}) x_v^tx_{v'}^t |
  \ &=\ Csh \label{eq:first_term}\\
%%%
\max_{v,v'}\ | \mathbb{E} [\sum_{t \in \mathcal{T}_n}
  w_t^\tau\eta(\xv^t; \thetav_u^{t}) x_v^tx_{v'}^t ] - q_{vv'}^\tau
|\ &=\ C'h. \label{eq:third_term}
\end{align}
%%%
Furthermore,
\begin{equation} \label{eq:second_term}
  | \sum_{t \in \mathcal{T}_n} 
    ( w_t^\tau\eta(\xv^t;\thetav_u^{t}) x_v^tx_{v'}^t - 
    \mathbb{E}[w_t^\tau\eta(\xv^t;\thetav_u^{t}) X_v^tX_{v'}^t])| 
   < \epsilon
\end{equation}
with probability at least $1 - 2 \exp ( -C''nh\epsilon^2 )$.
\end{lem}
%%%%
\begin{proof}
  We start the proof by bounding the difference $|\eta(\xv;
  \thetav_u^{t + \delta}) - \eta(\xv; \thetav_u^t)|$ which will be
  useful later on. By applying the mean value theorem to $\eta(\xv;
  \cdot)$ and the Taylor expansion on $\thetav_u^t$ we obtain:
%%%%%%%
\begin{equation*}
  \begin{aligned}
    | \eta(\xv;\thetav_u^{t+\delta}) - \eta(\xv;\thetav_u^{t})| & = 
     |\sum_{v=1}^{p-1} (\theta_{uv}^{t+\delta} - \theta_{uv}^{t})
      \eta'(\xv;\bar{\thetav}_u^{(v)}) | \qquad 
      \left( \begin{array}{c} 
        \bar{\thetav}_u^{(v)} \text{ is a point on the line} \\
        \text{between } \thetav_u^{t+\delta} \text{ and } \thetav_u^{t}\ 
      \end{array}\right)\\
  & \leq \sum_{v=1}^{p-1}|\theta_{uv}^{t+\delta} - \theta_{uv}^{t}| 
    \qquad (\ |\eta'(\xv; \cdot)| \leq 1\ )\\
  & = \sum_{v=1}^{p-1} |\delta \frac{\partial }{\partial t}\theta_{uv}^t +
      \frac{\delta^2}{2} \frac{\partial^2}{\partial t^2} 
           \theta_{uv}^t \Big|_{t=\beta_v}|
      \qquad \left( \begin{array}{c}
        \beta_v \text{ is a point on the line} \\
          \text{between  } t \text{ and } t + \delta
          \end{array} \right)
\end{aligned}
\end{equation*}
%%%%%%%

Without loss of generality, let $\tau = 1$. Using the above equation,
and the Riemann integral to approximate the sum, we have
%%%%
\begin{equation*} \label{eq:lem_pre_taylor}
  \begin{aligned}
    & | \sum_{t \in \mathcal{T}_n} 
         w_t^\tau \eta(\xv^t; \thetav_u^{\tau}) x_v^tx_{v'}^t -
      \sum_{t \in \mathcal{T}_n}
          w_t^\tau \eta(\xv^t; \thetav_u^{t}) x_v^tx_{v'}^t | \\
    & \approx |\int \frac{2}{h} K(\frac{z - \tau}{h}) [
       \eta(\xv^z;\thetav_u^{\tau}) - \eta(\xv^z;\thetav_u^{z}) ]
        x_v^zx_{v'}^z dz |\\
    & \leq 2\int_{-\frac{1}{h}}^{0} K(z') 
       | \eta(\xv^{\tau + z'h};\thetav_u^{ \tau}) - 
         \eta(\xv^{\tau + z'h};\thetav_u^{\tau + z'h})| dz'  \\
    & \leq 2\int_{-1}^{0} K(z') 
         [\sum_{v=1}^{p-1} |z'h \frac{\partial }{\partial t}
                \theta_{uv}^t\Big|_{t = \tau} + 
           \frac{(z'h)^2}{2} \frac{\partial^2 }{\partial t^2}
               \theta_{uv}^{t}\Big|_{t=\beta_v}|] dz' \\
    & \leq Csh,
\end{aligned}
\end{equation*}
for some constant $C > 0$ depending on $M$ from A3 which bounds the
derivatives in the equation above, and $M_K$ from A4 which bounds the
kernel. The last inequality follows from the assumption that the
number of non-zero components of the vector $\thetav_u^{t}$ is bounded
by $s$.

Next, we prove equation \eqref{eq:third_term}. Using the Taylor
expansion, for any fixed $ 1 \leq v,v' \leq p - 1$ we have
%%%
\begin{equation*}
 \begin{aligned}
  |\mathbb{E} & [\sum_{t \in \mathcal{T}_n}
      w_t^\tau\eta(\xv^t; \thetav_u^{t}) x_v^tx_{v'}^t ] - q_{vv'}^\tau | \\
  & = | \sum_{t \in \mathcal{T}_n} w_t^\tau(q_{vv'}^t - q_{vv'}^\tau) | \\
  & = | \sum_{t \in \mathcal{T}_n} w_t^\tau( 
         (t - \tau) \frac{\partial }{\partial t} q_{vv'}^t\Big|_{t =
           \tau} + 
       \frac{(t-\tau)^2}{2} \frac{\partial^2 }{\partial t^2} q_{vv'}^t
         \Big|_{t = \xi} |,
 \end{aligned}
\end{equation*}
%%%
where $\xi \in [t, \tau]$. Since $w_t^\tau = 0$ for $|t - \tau| > h$,
we have
\begin{equation*}
  \max_{v, v'}|\mathbb{E} [\sum_{t \in \mathcal{T}_n}
  w_t^\tau \eta(\xv^t; \thetav_u^{t}) x_v^tx_{v'}^t ] - q_{vv'}^\tau | \leq C'h\\
\end{equation*}
for some constant $C > 0$ depending on $M$ and $M_K$ only.

Finally, we prove equation \eqref{eq:second_term}. Observe that
$w_t^\tau\eta(\xv^t;\thetav_u^t) x_v^tx_{v'}^t$ are independent and
bounded random variables $[-w_t^\tau, w_t^\tau]$. The equation simply
follows from the Hoeffding's inequality.
\end{proof}

Using results of Lemma \ref{lem:large_dev:three_terms} we can obtain
the rate at which the element-wise distance between the true and
sample Fisher information matrix decays to zero as a function of the
bandwidth parameter $h$ and the size of neighborhood~$s$. In the proof
of the main theorem, the bandwidth parameter will be chosen so that
the bias and variance terms are balanced.

\subsection{Sample covariance matrix}

The deviation of the elements of the sample covariance matrix is
bounded in a similar way as the deviation of elements of the sample
Fisher information matrix, given in Lemma
\ref{lem:large_dev:three_terms}. Denoting the sample covariance matrix
at time point $\tau$ as
%%%%
\begin{equation*}
\hat \Sigmav^\tau = \sum_t w_t^\tau \xv^t \xv^{t'},
\end{equation*}
%%%%
and the difference between the elements of $\hat \Sigmav^\tau$ and
$\Sigmav^{\tau}$ can be bounded as 
%%%%
\begin{equation} \label{eq:large:covariance_decomposition}
\begin{aligned}
| \hat \sigma_{uv}^\tau - \sigma_{uv}^\tau | &\ =\ 
   | \sum_{t \in \mathcal{T}_n} w_t^\tau x_u^tx_v^t - \sigma_{uv}^\tau | \\
&\ \leq\ | \sum_{t \in \mathcal{T}_n} w_t^\tau x_u^tx_v^t - 
   \mathbb{E}[ \sum_{t \in \mathcal{T}_n} w_t^\tau x_u^tx_v^t ] | \\
&\ +\ | \mathbb{E}[ \sum_{t \in \mathcal{T}_n} w_t^\tau x_u^tx_v^t ] -
   \sigma_{uv}^\tau |.
\end{aligned}
\end{equation}
%%%%
The following lemma gives us bounds on the terms in
Eq.~\eqref{eq:large:covariance_decomposition}.
%%%%%%%%%%%%%%%%%%%%%%%%%%%%%%%%%%%%%%%%%%%%%%
%%%% begin lemma
%%%%%%%%%%%%%%%%%%%%%%%%%%%%%%%%%%%%%%%%%%%%%%
\begin{lem} \label{lem:covariance} Assume that the smoothness
  condition A3 is satisfied and that the kernel function $K(\cdot)$
  satisfies A4.  There are constants $C, C' > 0$ depending on $M$ and
  $M_K$ only such that for any $\tau \in [0,1]$, we have
%%%
\begin{equation} \label{eq:large_dev:cov:bias}
\max_{u,v} | \mathbb{E}[ \sum_{t \in \mathcal{T}_n} 
    w_t^\tau x_u^tx_v^t ] - \sigma_{uv}^\tau | \leq Ch.
\end{equation}
%%%
and 
%%%%
\begin{equation} \label{eq:large_dev:cov:variance}
| \sum_{t \in \mathcal{T}_n}  w_t^\tau x_u^tx_v^t - 
   \mathbb{E}[ \sum_{t \in \mathcal{T}_n}  w_t^\tau x_u^t x_v^t] | 
  \leq \epsilon 
\end{equation}
%%%%
with probability at least $1 -  2\exp(-C'nh\epsilon^2)$.
\end{lem}
%%%%
\begin{proof}
To obtain the Lemma, we follow the same proof strategy as in the proof
of Lemma~\ref{lem:large_dev:three_terms}. In particular,
Eq.~\eqref{eq:large_dev:cov:bias} is proved in the same way as
Eq.~\eqref{eq:third_term} and Eq.~\eqref{eq:large_dev:cov:variance} in
the same way as Eq.~\eqref{eq:second_term}. The details of this
derivation are omitted. 
\end{proof}

\section{Technical proofs}

In this appendix we provide proofs of lemmas used to prove the main
result.

\subsection{Proof of Lemma \ref{lem:same_non_zero}}

The set of minima $\Theta(\lambda_n)$ of a convex function is
convex. So, for two distinct points of minima, $\bar \thetav_u$ and
$\tilde{\thetav_u}$, every point on the line connecting two points
also belongs to minima, i.e. $\xi\bar{\thetav_u} +
(1-\xi)\tilde{\thetav_u} \in \Theta(\lambda_n)$, for any $\xi \in
(0,1)$. Let $\etav = \bar \thetav_u - \tilde{\thetav}_u$ and now any
point on the line can be written as $\tilde{\thetav}_u + \xi\etav$. The
value of the objective at any point of minima is constant and we have
%%%%
\begin{equation*}
F(\tilde{\thetav}_u + \xi\etav) = c,\quad \xi \in (0,1),
\end{equation*}
%%%%
where $c$ is some constant. By taking the derivative with respect to
$\xi$ of $F(\tilde{\thetav}_u + \xi\etav)$ we obtain
%%%%
\begin{equation} \label{eq:proof:first_derivative}
\begin{aligned}
\sum_{t \in \mathcal{T}_n} w_t^\tau &\left[
-x_u^t + 
\frac{\exp(\langle \tilde{\thetav}_u + \xi\etav, \xv_{\backslash u}^t
  \rangle) - 
\exp(-\langle \tilde{\thetav}_u + \xi\etav, \xv_{\backslash u}^t
  \rangle)}
{\exp(\langle \tilde{\thetav}_u + \xi\etav, \xv_{\backslash u}^t
  \rangle) + 
\exp(-\langle \tilde{\thetav}_u + \xi\etav, \xv_{\backslash u}^t
\rangle)}
\right]\langle \etav, \xv_{\backslash u}^t\rangle \\
&+ \lambda_n \sum_{v=1}^{p-1}\eta_v\sign(\tilde{\theta}_{uv} +
\xi\eta_v) = 0.
\end{aligned}
\end{equation}
%%%%
On a small neighborhood of $\xi$ the sign of $\tilde{\thetav}_u +
\xi\etav$ is constant,
%\mcomment{this holds without any smoothness conditions on $\thetav_u$}, 
 for each component $v$, since the
function $\tilde{\thetav}_u + \xi\etav$ is continuous in $\xi$. By
taking the derivative with respect to $\xi$ of
Eq.~\eqref{eq:proof:first_derivative} and noting that the last term is
constant on a small neighborhood of $\xi$ we have
%%%%
\begin{equation*}
4\sum_{t \in \mathcal{T}_n} w_t^\tau \langle \etav, \xv_{\backslash
  u}^t\rangle^2
\frac{\exp(-2\langle \tilde{\thetav}_u + \xi\etav, \xv_{\backslash u}^t
\rangle)}
{\left( 1 + \exp(-2\langle \tilde{\thetav}_u + \xi\etav, \xv_{\backslash u}^t
\rangle) \right)^2} = 0.
\end{equation*}
%%%%
This implies that $\langle \etav, \xv_{\backslash u}^t\rangle = 0$ for
every $t \in \mathcal{T}_n$, which implies that $\langle
\xv_{\backslash u}^t, \bar{\thetav}_u \rangle = \langle
\xv_{\backslash u}^t, \tilde{\thetav}_u \rangle$, $t \in {\cal T}_n$,
for any two solutions $\bar{\thetav}_u$ and $\tilde{\thetav}_u$. Since
$\bar{\thetav}_u$ and $\tilde{\thetav}_u$ were two arbitrary elements
of $\Theta(\lambda_n)$ we can conclude that $\langle \xv_{\backslash
  u}^t, \thetav_u \rangle$, $t \in \mathcal{T}_n$ is constant for all
elements $\thetav_u \in \Theta(\lambda_n)$.

Next, we need to show that the conclusion from above implies that any
two solutions have non-zero elements in the same position. From
equation \eqref{eq:kkt}, it follows that the set of non-zero
components of the solution is given by
%%%%
\begin{equation*}
S = \left\{ 1 \leq v \leq p - 1\ :\ \left|
\sum_{t \in \mathcal{T}_n}
w_t^\tau(\nabla \gamma(\thetav_u; \xv^t))_v\right| = \lambda \right\}.
\end{equation*}
%%%%
Using equation \eqref{eq:score} we have that 
%%%%
\begin{equation*}
\begin{aligned}
&\sum_{t \in \mathcal{T}_n} w_t^\tau(\nabla \gamma(\thetav_u^\tau; \xv^t))_v = \\ 
&\quad \sum_{t \in \mathcal{T}_n} w_t^\tau
(\xv_{\backslash u}^t \{ x_u^t + 1 - 
2\frac{ 
\exp (2x_u^{t}  \langle\thetav_u^\tau, \xv_{\backslash u}^t  \rangle ) } 
{ 
\exp (2x_u^{t}  \langle\thetav_u^{\tau},
\xv_{\backslash u}^{\tau}  \rangle ) + 1 }
\})_v,
\end{aligned}
\end{equation*}
%%%%
which is constant across different elements $\thetav_u \in
\Theta(\lambda_n)$, since $\langle \xv_{\backslash u}^t, \thetav_u
\rangle$, $t \in {\cal T}_n$ is constant for all $\thetav_u \in
\Theta(\lambda_n)$. This implies that the set of non-zero components
is the same for all solutions. \hfill $\Box$

\subsection{Proof of Lemma \ref{lem:proof:eigen_covariance}}

Under the assumptions given in the Lemma, we can apply the result of Lemma
\ref{lem:covariance}. Let $\yv \in
\mathbb{R}^{s}$ be a unit norm minimal eigenvector of $\hat
\Sigmav_{SS}^\tau$. We have
%%%%%
\begin{equation*}
\begin{aligned}
\Lambda_{\min}(\Sigmav_{SS}^\tau) &\ =\ 
\min_{\norm{\xv}_2 = 1} \xv'\Sigmav_{SS}^\tau \xv \\
&\ =\ \min_{\norm{\xv}_2 = 1}\ \{ 
\xv'\hat \Sigmav_{SS}^\tau \xv + 
\xv'(\Sigmav_{SS}^\tau - \hat \Sigmav_{SS}^\tau)\xv 
\ \}\\
&\ \leq\  \yv'\hat \Sigmav_{ SS}^\tau\yv +  
\yv'(\Sigmav_{ SS}^\tau - \hat \Sigmav_{ SS}^\tau)\yv,
\end{aligned} 
\end{equation*}
%%%%%
which implies 
%%%%%
\begin{equation*}
\Lambda_{\min}(\hat \Sigmav_{ SS}^\tau) \geq D_{\min} - 
\opnorm{(\Sigmav_{SS}^\tau - \hat \Sigmav_{SS}^\tau)}{2}.
\end{equation*}
%%%%%
Let $\Sigmav^\tau = [\sigma_{uv}^\tau]$ and $\hat
\Sigmav^\tau = [\hat \sigma_{uv}^\tau]$. We have the following
bound on the spectral norm
%%%%%
\begin{equation*}
\opnorm{\Sigmav_{SS}^\tau - \hat \Sigmav_{SS}^\tau}{2}
\leq \left( \sum_{u=1}^{s}\sum_{v=1}^{s} (\hat \sigma_{uv}^\tau -
\sigma_{uv}^\tau)^2 \right)^{1/2} \leq \delta,
\end{equation*}
%%%%%
with the probability at least $1 - 2\exp(-Cnh(\frac{\delta}{s} -
C'h )^2 + C''\log(s))$, for some fixed constants $C, C', C'' > 0$
depending on $M$ and $M_K$ only.

Similarly, we have that 
%%%%%
\begin{equation*}
\Lambda_{\max}(\hat \Sigmav_{SS}^\tau) \leq D_{\max} + \delta,
\end{equation*}
with probability at least $1 - 2\exp(-Cnh(\frac{\delta}{s} - C'h )^2 +
C''\log(s))$, for some fixed constants $C, C', C'' > 0$ depending on
$M$ and $M_K$ only.
%%%%%

From Lemma \ref{lem:same_non_zero}, we know that any two solutions
$\bar \thetav_u, \tilde \thetav_u \in \Theta(\lambda_n)$ of the
optimization problem \eqref{eq:opt_problem} have non-zero elements in
the same position. So, for any two solutions $\bar \thetav_u, \tilde
\thetav_u \in \Theta(\lambda_n)$, it holds
%%%%%
$$ \Xv_{\backslash u} (\bar \thetav_u - \tilde{\thetav}_u) =
\Xv_{\backslash u, S} (\bar \thetav_u - \tilde{\thetav}_u)_S +
\Xv_{\backslash u, S^c} (\bar \thetav_u - \tilde{\thetav}_u)_{S^c} = 
\Xv_{\backslash u, S} (\bar \thetav_u - \tilde{\thetav}_u)_S.
$$
%%%%%
Furthermore, from Lemma~\ref{lem:same_non_zero} we know that the two
solutions are in the kernel of $\Xv_{\backslash u, S}$. On the event
$\Omega_{01}$, kernel of $\Xv_{\backslash u, S}$ is $\{0\}$. Thus, the
solution is unique on $\Omega_{01}$.  \hfill $\Box$

\subsection{Proof of Lemma \ref{lem:proof:sample_fisher_information}}

We first analyze the probability of the event $\Omega_{02}$. Using the
same argument to those in the proof of
Lemma~\ref{lem:proof:eigen_covariance}, we obtain
%%%
  \begin{equation*}
    \Lambda_{\min}(\hat \Qv_{SS}^\tau) \geq C_{\min} - 
    \opnorm{\Qv_{SS}^\tau - \hat \Qv_{SS}^\tau}{2}.    
  \end{equation*}
%%%
  Next, using results of Lemma~\ref{lem:large_dev:three_terms}, we
  have the following bound
  \begin{equation} \label{eq:proof:eigen_fisher}
    \opnorm{\Qv_{SS}^\tau - \hat \Qv_{SS}^\tau}{2} \leq
    \left(\sum_{u=1}^{s}\sum_{v=1}^{s} (\hat q_{uv}^\tau -
      q_{uv}^\tau)^2 \right)^{1/2}\leq \delta,
  \end{equation}
  with probability at least $1 - 2\exp(-C\frac{nh\delta^2}{s^2} +
  2\log(s))$, for some fixed constants $C, C' > 0$ depending on $M$
  and $M_K$ only.

  Next, we deal with the event $\Omega_{03}$. We are going to use the
  following decomposition
  \begin{equation*}
    \begin{aligned}
      \hat \Qv_{S^cS}^\tau (\hat \Qv_{SS}^\tau)^{-1} &\ =\ \Qv_{S^cS}^\tau
      [(\hat \Qv_{SS}^\tau)^{-1} - (\Qv_{SS}^\tau)^{-1} ] \\
      &\ +\ [\hat \Qv_{S^cS}^\tau - \Qv_{S^cS}^\tau] (\Qv_{SS}^\tau)^{-1} \\
      &\ +\ [ \hat \Qv_{S^cS}^\tau - \Qv_{S^cS}^\tau ] 
      [(\hat \Qv_{SS}^\tau)^{-1} - (\Qv_{SS}^\tau )^{-1} ] \\
      &\ +\ \Qv_{S^cS}^\tau( \Qv_{SS}^\tau )^{-1} \\
      &\ =\ T_1 + T_2 + T_3 + T_4.
    \end{aligned}  
  \end{equation*}
  
  Under the assumption A2, we have that $\opnorm{T_4}{\infty} \leq 1 -
  \alpha$. The lemma follows if we prove that for all the other terms
  we have $\opnorm{\cdot}{\infty} \leq \frac{\alpha}{6}.$ Using the
  submultiplicative property of the norm, we have for the first term:
  \begin{equation} \label{eq:proof:t1}
    \begin{aligned}
      \opnorm{T_1}{\infty} &\ \leq\ 
        \opnorm{\Qv_{S^cS}^\tau \left( \Qv_{SS}^\tau \right)^{-1}}{\infty} 
        \opnorm{ \hat \Qv_{SS}^\tau - \Qv_{SS}^\tau}{\infty}
        \opnorm{ ( \hat \Qv_{SS}^\tau )^{-1}}{\infty} \\
      &\ \leq\ 
        (1 - \alpha) \opnorm{ \hat \Qv_{SS}^\tau - \Qv_{SS}^\tau}{\infty}
        \sqrt{s} \opnorm{ ( \hat \Qv_{SS}^\tau )^{-1}}{2}.
      \end{aligned}
  \end{equation}
  Using Eq.~\eqref{eq:proof:eigen_fisher}, we can bound the term
  $\opnorm{ \left( \hat \Qv_{SS}^\tau \right)^{-1}}{2} \leq C''$, for
  some constant depending on $C_{\min}$ only, with probability at
  least $1 - 2\exp(-C\frac{nh}{s} + 2\log(s))$, for some fixed
  constant $C > 0$. The bound on the term $\opnorm{ \hat \Qv_{SS}^\tau
    - \Qv_{SS}^\tau}{\infty}$ follows from application of Lemma
  \ref{lem:large_dev:three_terms}. Observe that
  \begin{equation}  \label{eq:proof:opnorm_fisher}
    \begin{aligned}
      \mathbb{P}[\opnorm{ \hat \Qv_{SS}^\tau - \Qv_{SS}^\tau}{\infty}
      \geq \delta]
      &\ = \ \mathbb{P}[\max_{v \in S} \{ \sum_{v' \in S}
      |\hat{q}_{vv'}^\tau -
      q_{vv'}^\tau| \} \geq \delta] \\
      & \ \leq \ 2\exp(-Cnh(\frac{\delta}{s} - C'sh)^2 +
      2\log(s)),
    \end{aligned}
  \end{equation}
  for some fixed constants $C, C' > 0$. Combining all the elements, we
  obtain the bound on the first term $\opnorm{T_1}{\infty} \leq
  \frac{\alpha}{6}$, with probability at least $1 -
  C\exp(C'\frac{nh}{s^3} + C''\log(s))$, for some
  constants $C, C', C'' > 0$.

  Next, we analyze the second term. We have that
  \begin{equation} \label{eq:proof:t2}
    \begin{aligned}
      \opnorm{T_2}{\infty} &\ \leq\ 
      \opnorm{\hat \Qv_{S^cS}^\tau -  \Qv_{S^cS}^\tau}{\infty} \sqrt{s} 
      \opnorm{\left( \Qv_{SS}^\tau \right)^{-1}}{2} \\
      &\ \leq\ 
      \frac{\sqrt{s}}{C_{\min}} 
      \opnorm{\hat \Qv_{S^cS}^\tau -  \Qv_{S^cS}^\tau}{\infty}.
    \end{aligned}    
  \end{equation}
  The bound on the term $\opnorm{ \hat \Qv_{SS}^\tau -
    \Qv_{SS}^\tau}{\infty}$ follows in the same way as the bound in
  Eq.~\eqref{eq:proof:opnorm_fisher} and we can conclude that
  $\opnorm{T_3}{\infty} \leq \frac{\alpha}{6}$ with probability at
  least $1 - C\exp(C'\frac{nh}{s^3} + C''\log(p))$, for some constants
  $C, C', C'' > 0$.

  Finally, we bound the third term $T_3$. We have the following
  decomposition
  \begin{equation*}
    \begin{aligned}
      \opnorm{[\hat{\Qv}_{S^cS}^\tau & -
        \Qv_{S^cS}^\tau][(\hat{\Qv}_{SS}^\tau)^{-1} -
      (\Qv_{SS}^\tau)^{-1}]}{\infty} \\
      &\ \leq \ 
      \opnorm{\hat{\Qv}_{S^cS}^\tau - \Qv_{S^cS}^\tau}{\infty} \sqrt{s} 
      \opnorm{(\Qv_{SS}^\tau)^{-1} 
        [ \Qv_{SS}^\tau -
        \hat{\Qv}_{SS}^\tau](\hat{\Qv}_{SS}^\tau)^{-1}}{2} \\
      & \ \leq \ 
      \frac{\sqrt{s}}{C_{\min}}
      \opnorm{\hat{\Qv}_{S^cS}^\tau - \Qv_{S^cS}^\tau}{\infty}
      \opnorm{\Qv_{SS}^\tau - \hat{\Qv}_{SS}^\tau}{2}
      \opnorm{(\hat{\Qv}_{SS}^\tau)^{-1}}{2}.      
    \end{aligned}
  \end{equation*}
  Bounding the remaining terms as in equations \eqref{eq:proof:t2},
  \eqref{eq:proof:opnorm_fisher} and \eqref{eq:proof:t1}, we obtain
  that $\opnorm{T_3}{\infty} \leq \frac{\alpha}{6}$ with probability
  at least $1 - C\exp(C'\frac{nh}{s^3} + C''\log(p))$. 

  Bound on the probability of event $\Omega_{03}$ follows from
  combining the bounds on all terms.
\hfill $\Box$

\subsection{Proof of Lemma~\ref{lem:converge_theta}}

To prove this Lemma, we use a technique of \citet{rothman08spice}
applied to the problem of consistency of the penalized covariance
matrix estimator. Let us define the following function
%%%%
\begin{equation*}
H : \left\{ \begin{array}{ccl}
\mathbb{R}^{p} & \rightarrow  & \mathbb{R} \\
\Dv & \mapsto &
F(\thetav^{\tau}_u + \Dv) - F(\thetav^{\tau}_u),
\end{array}
\right.
\end{equation*}
where the function $F(\cdot)$ is defined in equation
\eqref{eq:proof:def_F}. The function $H(\cdot)$ takes the following
form 
%%%
\begin{equation*}
  \begin{aligned}
    H(\Dv) = &\sum_{t \in \mathcal{T}_n}w_t^\tau
    (\gamma(\thetav^{\tau}_{u}; \xv^t)- \gamma(\thetav^{\tau}_{u} +
    \Dv; \xv^t)) \\
    &\ +\ \lambda_n (\norm{\thetav^{\tau}_{u} +
      \Dv}_1 - \norm{\thetav^{\tau}_{u}}_1).
  \end{aligned}
\end{equation*}
%%%

Recall the minimizer of \eqref{eq:opt_problem} constructed in the
proof of Proposition~\ref{prop:sign_consistency}, $\hat\thetav_u^\tau
= (\bar{\thetav}_S', 0_{S^c}')'$. The minimizer of the function
$H(\cdot)$ is $\hat \Dv = \hat\thetav_u^\tau - \thetav_{u}^\tau$.
Function $H(\cdot)$ is convex and $H(0) = 0$ by construction. Therefor
$H(\hat \Dv) \leq 0$. If we show that for some radius $B > 0$, and
$\Dv \in \mathbb{R}^{p}$ with $\norm{\Dv}_2 = B$ and $\Dv_{S^c} =
\mathbf{0}$, we have $H(\Dv) > 0$, then we claim that $\norm{\hat
  \Dv}_2 \leq B$. This follows from the convexity of $H(\cdot)$.

We proceed to show strict positivity of $H(\cdot)$ on the boundary of
the ball with radius $B = K\lambda_n\sqrt{s}$, where $K > 0$ is a
parameter to be chosen wisely later. Let $\Dv \in \mathbb{R}^{p}$ be
an arbitrary vector with $\norm{\Dv}_2 = B$ and $\Dv_{S^c} =
\mathbf{0}$, then by the Taylor expansion of $\gamma(\cdot; \xv^t)$ we
have
%%%%%
\begin{equation} \label{eq:def_G}
  \begin{aligned}
    H ( \Dv )&\ =\ 
     -(\sum_{t \in \mathcal{T}_n} w_t^\tau \nabla \gamma(\thetav^\tau_u;
     \xv^t))' \Dv \\
    &\quad\   - \Dv' [ 
        \sum_{t \in \mathcal{T}_n} 
          w_t^\tau \eta( \xv^t; \thetav_u^\tau + \alpha \Dv)   
        \xv_{\backslash u}^t\xv_{\backslash u}^{t'}] \Dv \\
    &\quad \ + 
     \lambda_n (\norm{\thetav_u^\tau + \Dv}_1 -
       \norm{\thetav_u^\tau}_1 )\\
    &\ =\  (I) + (II) + (III), \\
\end{aligned}
\end{equation}
%%%%%
for some $\alpha \in [0, 1]$. 

We start from the term $(I)$. Let $\ev_v \in \mathbb{R}^p$ be a unit
vector with one at the position $v$ and zeros elsewhere. Then random
variables $-\ev_v'\sum_{t \in \mathcal{T}_n} w_t^\tau \nabla
\gamma(\thetav^\tau_u; \xv^t)$ are bounded
$[-\frac{C}{nh},\frac{C}{nh}]$ for all $1 \leq v \leq p-1$, with
constant $C > 0$ depending on $M_K$ only. Using the Hoeffding
inequality and the union bound, we have
%%%%
\begin{equation*}
  \max_{1 \leq v \leq p-1} |\ev_v'(\sum_{t \in \mathcal{T}_n} 
    w_t^\tau \nabla \gamma(\thetav^\tau_u; \xv^t)  - 
   \mathbb{E}[\sum_{t \in \mathcal{T}_n} 
    w_t^\tau \nabla \gamma(\thetav^\tau_u; \xv^t)])| \leq \delta,
\end{equation*}
%%%%
with probability at  least $1-2\exp(-Cnh\delta^2 + \log(p))$, where
$C >  0$ is  a constant depending  on $M_K$ only.  Moreover, denoting
%%%
$$
p(\thetav_u^t) = \mathbb{P}_{\thetav_u^t} [ x_u^t = 1\ |\
\xv_{\backslash u}^t ]
$$
%%% 
to simplify the notation, we have for all $1 \leq v \leq p-1$,
%%%%
\begin{equation}
  \begin{aligned} \label{eq:expectation_w}
    |\mathbb{E} & [ \ev_v' \sum_{t \in \mathcal{T}_n} w_t^\tau 
      \nabla \gamma(\thetav^\tau_u; \xv^t)\ |\ 
      \{ \xv_{\backslash u}^t \}_{t \in \mathcal{T}_n} ]| \\
    &\ =\ | \mathbb{E} [ \sum_{t \in \mathcal{T}_n} w_t^\tau
         x_v^t [ x_u^t + 1 - 2p(\thetav_u^\tau) ]\ |\ 
      \{ \xv_{\backslash u}^t \}_{t \in \mathcal{T}_n} ] | \\
    &\ =\ | 2 \sum_{t \in \mathcal{T}_n} w_t^\tau
         x_v^t [  p(\thetav_u^{t}) - p(\thetav_{u}^{\tau}) ] | \\
    &\ \leq\ 4 \int_{-\frac{1}{h}}^{0} K(z) 
        | p ( \thetav_u^{\tau + zh} ) -
          p (\thetav_{u}^{ \tau}) |  dz. \\
\end{aligned}
\end{equation}
%%%%
Next, we apply the mean value theorem on $p(\cdot)$ and the Taylor's
theorem on $\thetav_u^t$. Under the assumption A3, we have
\begin{equation} \label{eq:app:prob}
  \begin{aligned}
    & | p ( \thetav_u^{\tau + zh} ) - p (\thetav_u^{ \tau}) | \\
    &\qquad \leq \ \sum_{v=1}^{p-1} | \thetav_{uv}^{\tau + zh} -
    \theta_{uv}^{ \tau}|\qquad( |\ p'(\cdot)| \leq 1\ ) \\
    &\qquad =\ \sum_{v=1}^{p-1} 
       | zh \frac{\partial}{\partial t} \theta_{uv}^t \Big|_{t =\tau} 
       + \frac{(zh)^2}{2} \frac{\partial^2}{\partial t^2} 
         \theta_{uv}^t\Big|_{t = \alpha_v}  |
       \qquad (\ \alpha_v \in [\tau + zh, \tau] \ ) \\
    &\qquad \leq\ Cs |zh + \frac{(zh)^2}{2}|,
\end{aligned}
\end{equation}
for some $C > 0$ depending only on $M$. Combining \eqref{eq:app:prob}
and \eqref{eq:expectation_w} we have that $|\mathbb{E}[ \ev_v' \sum_{t
  \in \mathcal{T}_n} w_t^\tau \nabla \gamma(\thetav^\tau_u; \xv^t)|
\leq Csh$ for all $1 \leq v \leq p-1$. Thus, with probability greater
than $1 - 2\exp(-Cnh(\lambda_n - sh)^2 + \log(p))$ for some constant
$C > 0$ depending only on $M_K, M$ and $\alpha$, which under the
conditions of Theorem \ref{thm:main} goes to 1 exponentially fast, we
have
%%%%
\begin{equation*}
  \max_{1 \leq v \leq p-1} |\ev_v'\sum_{t \in \mathcal{T}_n} 
     w_t^\tau \nabla \gamma(\thetav^\tau_u; \xv^t)| \leq 
  \frac{\alpha\lambda_n}{4(2-\alpha)} < \frac{\lambda_n}{4}.
\end{equation*}
%%%%
On that event, using H\"{o}lder's inequality, we have
%%%%
\begin{equation*}
  \begin{aligned}   
    |(\sum_{t \in \mathcal{T}_n} w_t^\tau \nabla \gamma(\thetav^\tau_u;
      \xv^t))' \Dv| & 
    \leq \norm{\Dv}_1 \max_{1 \leq v \leq p-1} |\ev_v'\sum_{t
      \in \mathcal{T}_n} w_t^\tau \nabla \gamma(\thetav^\tau_u; \xv^t)|
       \\ &
    \leq \frac{\lambda_n}{4} \sqrt{s} \norm{\Dv}_2
      \leq (\lambda_n\sqrt{s})^2\frac{K}{4}.
  \end{aligned}
\end{equation*}
%%%%
The triangle inequality applied to the term $(III)$ of equation
\eqref{eq:def_G} yields:
%%%
\begin{equation*}
  \begin{aligned}
    \lambda_n (\norm{\thetav_u^\tau + \Dv}_1 -
      \norm{\thetav_u^\tau}_1) 
      & \geq -\lambda_n \norm{\Dv_S}_1 \\
    & \geq -\lambda_n \sqrt{s}
      \norm{\Dv_S}_2 \geq -K (\lambda_n\sqrt{s})^2.
  \end{aligned}
\end{equation*}
%%%
Finally, we bound the term $(II)$ of equation
\eqref{eq:def_G}. Observe that since $\Dv_{S^c} = 0$, we
have
%%%%
\begin{equation*}
  \begin{aligned}
    \Dv' &[ \sum_{t \in \mathcal{T}_n} w_t^\tau \eta( \xv^t;
      \thetav_u^\tau + \alpha \Dv) \xv_{\backslash u}^t\xv_{\backslash u}^{t'}]
      \Dv \\
    &\ =\ \Dv_S' [ \sum_{t \in \mathcal{T}_n} w_t^\tau
       \eta( \xv^t; \thetav_u^\tau + \alpha \Dv) 
       \xv_{S}^t\xv_{S}^{t'}] \Dv_S \\
    &\ \geq \ K^2 \Lambda_{\min}( \sum_{t \in \mathcal{T}_n}
       w_t^\tau \eta(\xv^t; \thetav_{u}^\tau + \alpha \Dv) 
       \xv_S^t\xv_S^{t'} )
  \end{aligned}
\end{equation*}
%%%%
Let $g : \mathbb{R} \mapsto \mathbb{R}$ be defined as $g(z) = \frac{4
  \exp(2z)}{(1 + \exp(2z))^2}$. Now, $\eta(\xv; \thetav_u) = g(x_u \langle
\thetav_u, \xv_{\backslash u} \rangle)$ and we have
\begin{equation*}
  \begin{aligned}
    \Lambda_{\min} &( \sum_{t \in \mathcal{T}_n}
       w_t^\tau\eta(\xv^t; \thetav_u^\tau + \alpha \Dv) 
       \xv_S^t\xv_S^{t'} ) \\
    &\ \geq\ \min_{\alpha \in [0,1]}
       \Lambda_{\min} ( \sum_{t \in \mathcal{T}_n} w_t
       \eta( \xv^t; \thetav_u^\tau + \alpha \Dv) 
       \xv_S^t \xv_S^{t'}) \\
    &\ \geq\ \Lambda_{\min} ( \sum_{t \in \mathcal{T}_n} w_t^\tau
         \eta( \xv^t; \thetav^\tau_u )
      \xv_S^t\xv_S^{t'}) \\
    & \qquad - \max_{\alpha \in [0,1]} 
         \opnorm{\sum_{t \in \mathcal{T}_n} w_t^\tau
         g'(x_u^t \langle \thetav_u^\tau + \alpha
            \Dv, \xv_S^t \rangle) 
         (x_u^t \Dv_S' \xv_S^t) \xv_S^t\xv_S^{t'}}{2} \\
    &\ \geq\ C_{\min} - \max_{\alpha \in [0,1]} 
       \opnorm{\sum_{t \in \mathcal{T}_n} w_t^\tau
         g'( x_u^t \langle \thetav_u^\tau + \alpha
           \Dv, \xv_S^t \rangle) 
        (x_u^t \Dv_S' \xv_S^t) \xv_S^t \xv_S^{t'}}{2} \\
\end{aligned}
\end{equation*}
To bound the spectral norm, we observe that for any fixed $\alpha \in
[0,1]$ and $y \in \mathbb{R}^{s}, \norm{\yv}_2 = 1$ we have:
\begin{equation*}
  \begin{aligned}
    \yv'& \{ \sum_{t \in \mathcal{T}_n} w_t^\tau
      g'(x_u^t\langle\thetav_u^\tau + \alpha\Dv, 
         \xv_S^t\rangle) (x_u^t \Dv_S' \xv_S^t) 
      \xv_S^t \xv_S^{t'} \} \yv\\
    & = \sum_{t \in \mathcal{T}_n} w_t^\tau
      g'(x_u^t\langle\thetav_u^\tau + \alpha\Dv, 
         \xv_S^t \rangle ) (x_u^t \Dv_S' \xv_S^t )
         ( \xv_S^{t'}\yv )^2 \\
    & \leq \sum_{t \in \mathcal{T}_n} w_t^\tau| 
      g'(x_u^t\langle\thetav_u^\tau + \alpha\Dv, 
         \xv_S^t \rangle ) (x_u^t \Dv_S' \xv_S^t )|
         ( \xv_S^{t'}\yv )^2 \\
    & \leq \sqrt{s}\norm{\Dv}_2 
         \opnorm{\sum_t w_t^\tau \xv_S^t\xv_S^{t'}}{2}
       \qquad(\ |g'(\cdot)| \leq 1\ ) \\
    & \leq D_{\max} K\lambda_ns 
     \leq \frac{C_{\min}}{2}.
  \end{aligned}
\end{equation*}
The last inequality follows as long as $\lambda_ns \leq
\frac{C_{\min}}{2D_{\max}K}$. We have shown that
%%%%
\begin{equation*}
  \Lambda_{\min} ( \sum_{t \in \mathcal{T}_n}
       w_t^\tau\eta(\xv^t; \thetav_u^\tau + \alpha \Dv) 
       \xv_S^t\xv_S^{t'} ) \geq \frac{C_{\min}}{2},
\end{equation*}
%%%
with high probability. 

Putting the bounds on the three terms together, we have
%%%
\begin{equation*}
  H(\Dv)\ \geq\ (\lambda_n\sqrt{s})^2 \left\{
    -\frac{1}{4}K + \frac{C_{\min}}{2}K^2 - K \right\},
\end{equation*}
%%%
which is strictly positive for $K = \frac{5}{C_{\min}}$. For this
choice of $K$, we have that $\lambda_n s \leq \frac{ C_{\min}^2 }
{10 D_{\max}}$, which holds under the conditions of
Theorem~\ref{thm:main} for $n$ large enough.
\hfill $\Box$

\bibliography{biblio}

\end{document}